\documentclass{article}

\usepackage{arxiv}

\usepackage[utf8]{inputenc}
\usepackage[T1]{fontenc}
\usepackage{url}
\usepackage{booktabs}
\usepackage{amsfonts}
\usepackage{nicefrac}
\usepackage{microtype}
\usepackage{graphicx}
\usepackage{natbib}
\usepackage{doi}
\usepackage{comment}
\usepackage{csquotes}
\usepackage{subcaption}
\usepackage{enumitem}
\usepackage[table,dvipsnames]{xcolor}
\usepackage{colortbl}
\usepackage{soul}
\usepackage{placeins}
\usepackage{tabularx}
\usepackage{array}
\usepackage{float}
\usepackage{adjustbox}
\usepackage{lscape}
\usepackage{hyperref}  

\usepackage{tcolorbox}

\definecolor{steel}{RGB}{70,130,180}  
\definecolor{todocolor}{RGB}{0, 70, 180}
\tcbuselibrary{breakable}

\title{Automated Clinical Report Generation for Remote Cognitive Remediation: Comparing Knowledge-Engineered Templates and LLMs in Low-Resource Settings}

\author{ 
	{Yongxin ZHOU, Fabien RINGEVAL, and François PORTET} \\
Univ. Grenoble Alpes, CNRS, Grenoble INP, LIG, 38000 Grenoble, France \\
\texttt{firstname.lastname@univ-grenoble-alpes.fr} \\
}

\date{}

\renewcommand{\shorttitle}{Automated Clinical Report Generation for Remote Cognitive Remediation}

\hypersetup{
pdftitle={Automated Clinical Report Generation for Remote Cognitive Remediation: Comparing Knowledge-Engineered Templates and LLMs in Low-Resource Settings},
pdfauthor={Yongxin ZHOU, Fabien RINGEVAL, and François PORTET},
pdfkeywords={Natural Language Generation, Clinical Report Generation, Knowledge Engineering, Large Language Models, Digital Cognitive Therapy, Human Evaluation},
}

\begin{document}
\maketitle

\begin{abstract}
The growing demand for cognitive remediation therapy, combined with limited speech therapist availability, has accelerated the adoption of remote and semi-autonomous rehabilitation tools. These systems generate large volumes of interaction data that are difficult for clinicians to review efficiently. This paper investigates automated clinical report generation for avatar-guided, home-based cognitive remediation sessions, addressing a low-resource setting in which no reference reports exist.
We present and compare two complementary approaches: (1) a rule-based template system that constitutes a clinical expert system in the classical sense, encoding speech therapy domain knowledge as explicit decision rules and validated report templates, and designed to ensure clinical reliability and traceability; and (2) a zero-shot LLM-based approach (GPT-4) aimed at producing more fluent and concise output. Both systems take identical pre-extracted, expert-validated structured variables as input, enabling a controlled factual comparison. Outputs were evaluated by eight certified speech therapists and final-year speech therapy students using a nine-criterion structured questionnaire.
Results reveal a clear trade-off between clinical reliability and linguistic quality. The template-based system scored higher on fluidity, coherence, and results presentation, while GPT-4 produced more concise output. Directional differences are consistent across all evaluation dimensions, though no comparison reached statistical significance after correction, reflecting the inherent scale constraints of expert clinical evaluation.
Based on systematic analysis of evaluator feedback, we derive eight design recommendations for clinical report generation systems in remote rehabilitation settings. More broadly, this work contributes a replicable methodology for building and assessing clinical NLG systems in low-resource settings: expert elicitation, taxonomy-driven generation, and multi-dimensional human evaluation. It also illustrates how controlled comparisons can inform the responsible adoption of generative AI in healthcare.\footnote{Code and data available at \url{https://github.com/yongxin2020/remediation-report-generator}.}
\footnote{Under review.}
\end{abstract}


\section{Introduction}
\label{sec:intro}

The evolution of healthcare towards predictive, preventive, personalized, and participatory care (P4 medicine) \citep{P4_medicine2013}, combined with advances in digital technology and AI, has facilitated the development of non-drug treatments such as digital therapies. One prominent example is \textit{\textbf{cognitive remediation}}, a form of therapy that provides interactive and configurable stimulation exercises to individuals with cognitive conditions \citep{cognitive_smartphone}.

Cognitive remediation interventions targeting cognitive decline encompass three main subtypes: \textit{cognitive training} (CT), \textit{cognitive rehabilitation} (CR), and \textit{cognitive stimulation} (CS) \citep{tulliani2022efficacy}. This study focuses on CT, a restorative approach aimed at improving cognitive performance through structured tasks \citep{BaharFuchs2019}, employing techniques such as memory strategies (e.g., cueing, method of loci) and repetitive exercises (e.g., spaced retrieval training, attention and recall tasks) to strengthen specific cognitive domains \citep{Mowszowski_Batchelor_Naismith_2010}.

In clinical practice, therapists, typically speech therapists or neuropsychologists in France, design therapeutic plans around face-to-face sessions with patients. Session frequency is often constrained by therapist availability, and while intensive home-based training could improve treatment efficacy, patients frequently struggle to adhere to independent exercises \citep{Turunen2019}. Effective solutions must therefore both support patient engagement in autonomous remediation and preserve therapist oversight.

\begin{figure*}
    \centering
    \includegraphics[width=.9\textwidth]{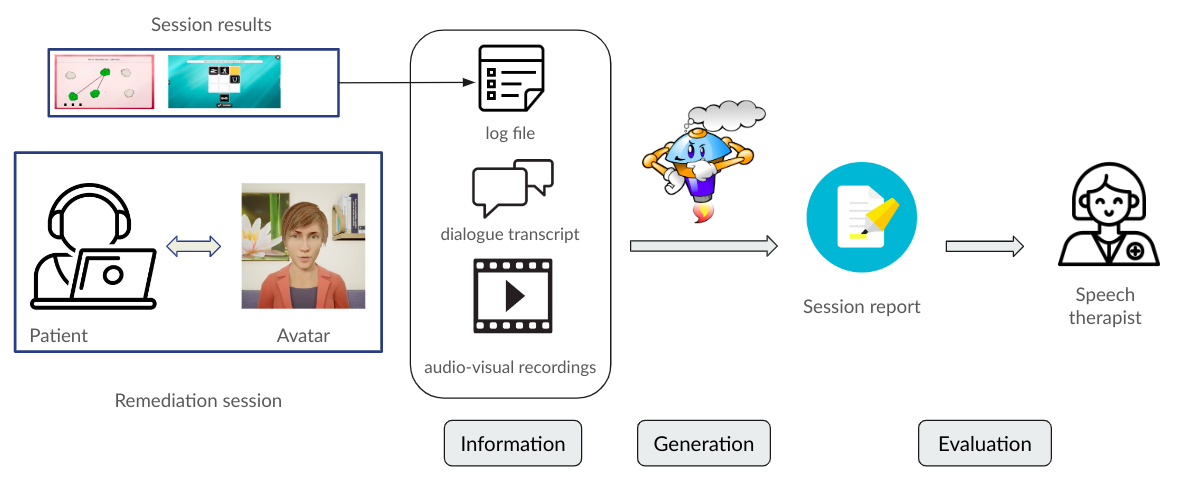}
    \caption{Framework for summarizing remediation sessions in THERADIA. The system processes two data types: (1) high-level session results (LOG files) and (2) low-level raw data (e.g., conversation transcripts in CSV files, audio-visual recordings). Three research questions guide report generation: (1) content selection (What to include?), (2) system design (How to generate?), and (3) evaluation (How to assess quality?).}
    \label{fig:theradia_nlg}
\end{figure*}

The THERADIA project \citep{10.1007/978-3-030-80285-1_55} addresses this challenge by deploying a virtual assistant to guide patients through home-based remediation sessions. Our work operates within this framework, focusing on the automatic generation of clinical reports that summarize these sessions to support therapist follow-up. As illustrated in Figure~\ref{fig:theradia_nlg}, sessions produce two types of data: high-level results (\textit{LOG} files) and low-level raw data (conversation transcripts, audiovisual recordings). Converting this heterogeneous data into actionable clinical reports raises three research questions:

\begin{enumerate}[noitemsep,topsep=2pt]
    \item \textbf{Content Selection}: Given the breadth of available session data, what information should clinical reports include for speech therapists? This involves: (1) how to reliably capture and quantify domain-specific variables, and (2) how to prioritize report content according to clinical needs.
    \item \textbf{System Design}: Given the requirements for accuracy, controllability, and fluency in clinical contexts, and the absence of reference reports in this low-resource setting; what generation approach is most appropriate: an expert-driven rule-based pipeline, or an LLM-based approach?
    \item \textbf{Evaluation}: What constitutes an appropriate method for assessing the quality of generated reports in this clinical context?
\end{enumerate}

To address these questions, we combined data-driven and expert-based methods, working closely with practicing speech therapists through regular meetings to identify reporting needs and assess data constraints.

\paragraph{Contributions} The main contributions of this work are:

\begin{itemize}[noitemsep,topsep=2pt]
    \item A \textbf{domain taxonomy} of clinically relevant reporting elements for home-based cognitive remediation, developed through iterative collaboration with practicing speech therapists.
    \item A \textbf{rule-based template generation system} that embodies the classical expert system paradigm, encoding speech therapy domain knowledge as explicit decision rules and validated templates, producing structured, traceable clinical reports.
    \item A \textbf{comparative evaluation} of template-based and LLM-based (GPT-4) report generation, assessed by certified speech therapists through quantitative scoring and structured qualitative feedback.
    \item \textbf{Evidence-based design recommendations} for clinical NLG systems in low-resource rehabilitation settings, grounded in systematic expert feedback analysis.
\end{itemize}

\section{Related Work}
\label{sec:related_work}

\subsection{Clinical NLP and Automated Medical Report Generation}

The automatic generation of clinical text from patient data has been an active area of NLG research for over two decades \citep{Reiter2000}, progressing from patient-tailored written information \citep{REITER200341} to neonatal intensive care summaries \citep{PORTET2009789} and ICU shift handover notes \citep{HUNTER2012157}. Persistent challenges include the scarcity of reference texts in specialized domains, heterogeneous documentation standards across institutions, and strict factual accuracy requirements in safety-critical settings \citep{zhou-etal-2023-survey}.

A key distinction within clinical NLG is between \textit{narrative report generation}, where systems describe events or processes in natural prose, and \textit{data-to-text generation}, which transforms structured or semi-structured data (e.g., vital signs, test scores, interaction logs) into readable summaries \citep{10.5555/3241691.3241693, Pauws2019}. Our work falls in the latter category: exercise outcomes, linguistic indicators, and affective states are encoded as structured variables and rendered as a clinical narrative. This framing is common in clinical monitoring systems, where the primary objective is to communicate objective measurements clearly to therapists or caregivers \citep{zhou:tel-05101086}.

An additional challenge specific to our setting is the \textit{low-resource} nature of the domain: no reference reports exist, and the small number of recorded sessions precludes supervised training.
This configuration calls for either expert-designed generation rules or zero-shot prompting of pre-trained language models, the two complementary approaches compared in this work.

\subsection{Template-based and Rule-based NLG}

Template-based and rule-based approaches have a long and successful history in clinical NLG, precisely because their outputs are transparent, auditable, and free of hallucination, all properties essential in safety-critical healthcare contexts \citep{Reiter2000}. 
Landmark systems include \textsc{SumTime} \citep{reiter-sripada-2003-learning}, which generates weather forecasts as natural-language text; \textsc{SemScribe} \citep{varges-etal-2012-semscribe}, which produces cardiological doctor's letters via configurable document plans and sentence frames; \textsc{STOP} \citep{REITER200341}, which generates personalized smoking-cessation letters; and \textsc{BabyTalk} \citep{PORTET2009789}, which produces clinical summaries from neonatal intensive care unit data streams. 
The latter is particularly relevant, as it demonstrates the feasibility of producing clinically validated narrative summaries from structured monitoring data through iterative co-design with domain experts. This iterative design tradition is equally central to \citet{REITER200341}, where clinician feedback substantially reshaped content selection and document structure across multiple evaluation cycles.

A common design pattern across these systems, and one we adopt here, combines an expert-validated content taxonomy with template sentences populated at runtime with computed variable values \citep{10.5555/3241691.3241693}. This approach affords full control over output content, prevents unsupported clinical inferences, and ensures structural consistency across reports. Its primary limitation is the tendency to produce formulaic text that can appear repetitive when the same template structure recurs across reports \citep{Reiter2000}, a trade-off that motivates the exploration of generative alternatives. Our methodology follows this established tradition, with practicing speech therapists involved throughout taxonomy development, template design, and output evaluation.

\subsection{Large Language Models for Clinical Text Generation}
Instruction-tuned LLMs have shown strong zero-shot generalization \citep{10.1145/3777411, zhou-etal-2025-gpt}, making them attractive for clinical text generation without task-specific training data. Recent work evaluates LLMs across diverse clinical tasks \citep{Bedi2026}, demonstrating their potential for a range of documentation applications, including radiology report structuring \citep{YANG2023100007}, biomedical signal summarization \citep{Liu2023.06.28.23291916}, and clinical note generation \citep{wang-etal-2025-towards-adapting}. These studies highlight the capacity of LLMs to interpret complex inputs and produce coherent prose, making them promising candidates for automating clinical documentation workflows. 

However, deploying LLMs in clinical settings raises significant safety concerns. Hallucination, the generation of plausible but factually unsupported content, poses a fundamental risk in contexts requiring strict factual fidelity \citep{Thirunavukarasu2023Large, kalai2025languagemodelshallucinate}. Even highly capable medical LLMs can produce confident but clinically incorrect statements \citep{Singhal2023Large}.
In clinical documentation, a fabricated exercise outcome or inferred emotional state could lead to inappropriate therapeutic decisions, a risk compounded in low-resource settings where fine-tuning is infeasible.

\subsection{Digital Cognitive Rehabilitation and Remote Monitoring}
Computerized cognitive training (CCT) has been extensively studied for maintaining or improving cognitive function in older adults and individuals with mild cognitive impairment \citep{kueider-etal-2012-cct, BaharFuchs2019}. Meta-analyses confirm modest but consistent effects, with higher training intensity and session frequency producing larger gains \citep{Lampit2014Computerized, tulliani2022efficacy}. These findings have motivated home-based and remote CCT platforms that enable more frequent practice without in-person therapist presence \citep{Turunen2019}.

Remote deployment, however, introduces a critical clinical challenge: therapists lose direct observational access to patient behavior during sessions. Commercial platforms such as HappyNeuron\footnote{\url{https://www.happyneuron.fr/}} and BrainHQ\footnote{\url{https://www.brainhq.com/}} provide aggregate performance statistics, but these coarse summaries omit the qualitative behavioral, linguistic, and emotional information therapists rely on for individualized planning. The THERADIA project \citep{10.1007/978-3-030-80285-1_55} addresses this gap via a virtual assistant that guides patients through CCT exercises at home, generating rich multimodal session data: video, audio, transcripts, and structured logs. Our work builds on this platform, tackling the downstream task of converting raw session data into structured clinical reports, which, to our knowledge, has not been previously addressed in the CCT monitoring literature.

\subsection{Evaluation of Automatically Generated Clinical Text}

Evaluating automatically generated clinical text is particularly challenging in low-resource settings, where reference reports are unavailable and automated string-overlap metrics such as BLEU or ROUGE are therefore inapplicable \citep{zhou-etal-2023-survey}.
Human evaluation by certified domain experts is consequently the standard and most reliable approach, despite its cost and limited scalability.

Structured human evaluation frameworks have been proposed for a range of clinical and consultative NLG tasks.
\citet{savkov-etal-2022-consultation} develop a questionnaire-based evaluation for consultation note generation that assesses factual accuracy, completeness, and clinical utility, which we adapt for our setting.
\citet{van-der-lee-etal-2019-best} provide best-practice guidelines for human NLG evaluation, recommending clearly defined rating criteria, multi-dimensional assessment, and sufficient evaluator expertise — all principles reflected in our evaluation design.
The use of Likert-scale questionnaires administered to certified practitioners is consistent with evaluation practice in applied clinical NLG \citep{PORTET2009789, LYU2026104997}.

A methodological challenge specific to our context is the \textit{subjective} nature of clinical report quality: practicing speech therapists have markedly different preferences regarding report content, length, and structure.
This variability is well-documented in clinical NLG evaluation and argues for instruments that separate objective dimensions (factual accuracy, completeness) from stylistic ones (fluency, conciseness).
Our nine-criterion questionnaire reflects this distinction and provides a foundation for future comparative evaluations in this domain.

\section{Designing Remediation Session Reports: Data, Structure, and Key Information}
\label{sec:theradia_report_form}

\subsection{Data Collection and THERADIA Corpus}
\label{subsec:theradia_data}

The THERADIA project aims to develop a virtual assistant to support patients during home-based therapy sessions. To evaluate the concept and collect realistic data, the team conducted a Wizard-of-Oz experiment \citep{10.1007/978-3-030-80285-1_55}, in which participants interacted with a simulated embodied conversational agent. Eight exercises targeting various cognitive functions were proposed during these sessions (see Appendix~\ref{appendix:theradia_exercises}).

\begin{figure*}
    \centering
    \includegraphics[width=.9\textwidth]{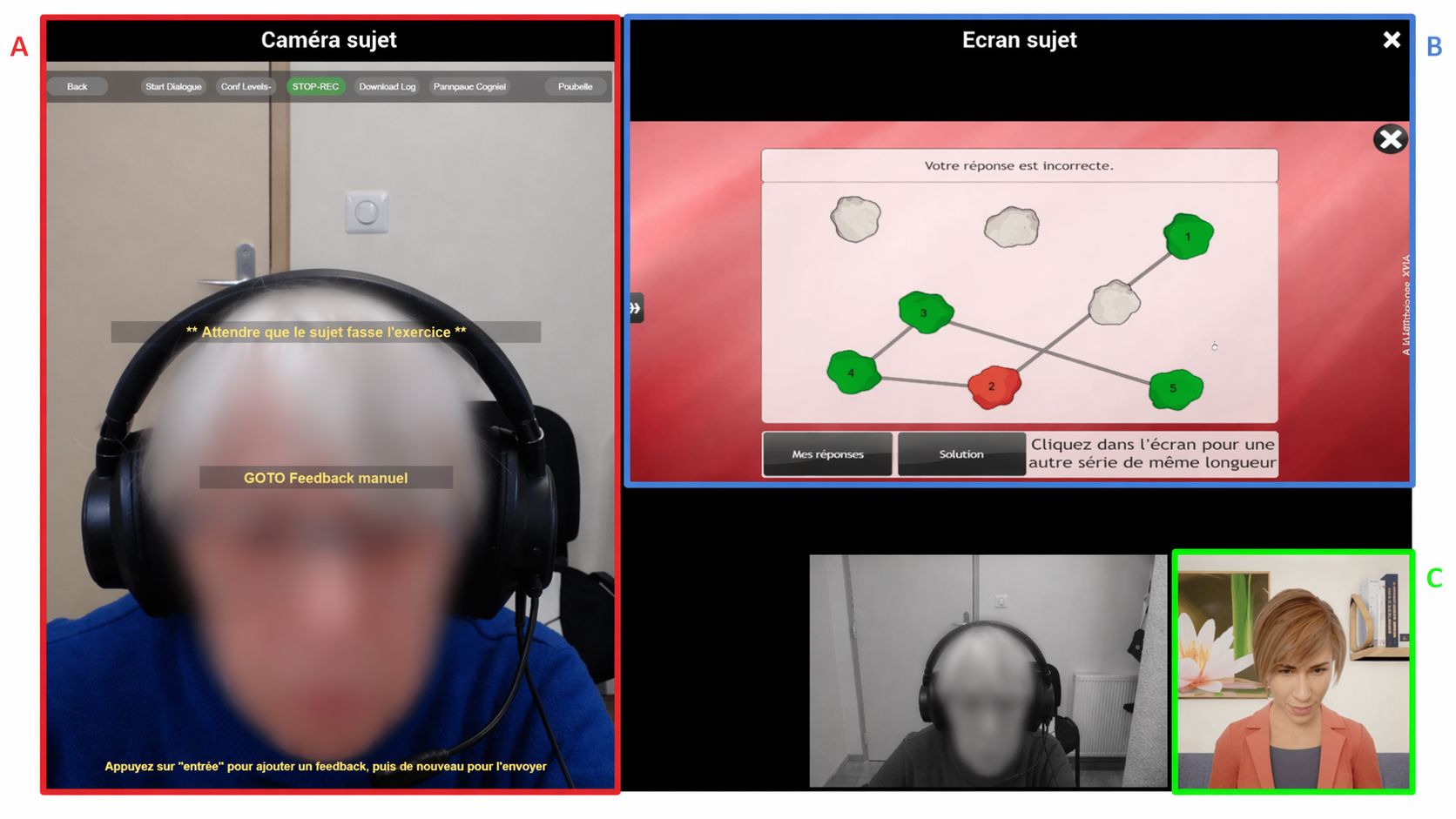}
    \caption{Screenshot of the video recorded during a remediation session between a participant and the avatar (operated by a human in the Wizard-of-Oz experiment). (A) participant's video feed, (B) participant's screen display, (C) virtual assistant rendering, displayed throughout the session.}
    \label{fig:video_session_capture}
\end{figure*}

\begin{table*}[ht]
    \centering
    \footnotesize
    \caption{Excerpt (translated from French) from the conversation between a participant and the virtual assistant during the Wizard-of-Oz experiment (from \textit{A01A\_1\_seg\_suzy\_subject}). \textit{Subject} refers to the participant; \textit{Suzy} refers to the avatar.}
    \begin{tabular}{lp{11cm}}
\toprule
\textbf{Speaker} & \textbf{Dialogue} \\
\midrule
\multicolumn{2}{c}{exo1: successful exercise} \\
\midrule
Suzy  	& You have just completed your first exercise! How did it go? \\
Subject	& I think not too bad, I'd even say almost good \\
Suzy & That's true, it was very good! \\
Suzy	& Did you use a specific strategy to succeed? \\
Subject	& no, just observation \\
\midrule
\multicolumn{2}{c}{exo5: failed exercise} \\
\midrule
Suzy & You have just finished this exercise for the first time, it was a bit difficult, wasn't it? \\
Subject & uh yes and I think I'm lacking concentration right now \\
Suzy & Maybe you tried to answer too quickly and didn't have time to see certain numbers? Is that what caused you problems? \\
Subject & maybe \\
\bottomrule
\end{tabular}
\label{tab:theradia_dialogue_example}
\end{table*}

In this scenario, the patient performs exercises alone at home while the virtual assistant guides them through the session. A screenshot of the recording setup is shown in Figure~\ref{fig:video_session_capture}, and a conversation excerpt is provided in Table~\ref{tab:theradia_dialogue_example}, illustrating exchanges following both a successful and a failed exercise. Log files record timestamps, Wizard button presses (\textit{LOG|REP}), predefined avatar dialogue (\textit{LOG|TXT}), and exercise outcomes (\textit{LOG|ENDGAME}); an excerpt is provided in Appendix~\ref{fig:log_excerpt}.

The corpus comprises 52 young participants ($<$30 years old), 52 seniors ($\geq$65 years old), and 9 participants with Mild Cognitive Impairment (MCI) associated with early-stage Alzheimer's disease (data statistics in Appendix~\ref{appendix:theradia_statistics}). Middle-aged individuals were excluded as the rehabilitation program was designed specifically for older adults with age-related cognitive difficulties. Young and senior participants completed one or two sessions, each comprising 8 exercises (one repetition each) lasting 1–1.5 hours. MCI participants completed one session with four exercises (Exo1, 2, 3, and 7), each repeated twice (8 activities total), lasting 30–45 minutes. Each recorded session yielded video recordings (Figure~\ref{fig:video_session_capture}), conversation transcripts (Table~\ref{tab:theradia_dialogue_example}), and log files (Figure~\ref{fig:log_excerpt}).

\subsection{Information to be Presented in a Remediation Report}
\label{subsubsec:remote_variables}

Through regular quarterly meetings with two speech therapists from the THERADIA project, we established that clinical reports for home-based sessions should be equivalent to a handover note written by a substitute therapist: comprehensible to the receiving therapist and covering all major session information.

These consultations led to the identification of essential elements for remote clinical observation, proposed by an expert collaborating speech therapist based on her clinical experience (detailed in Table~\ref{table:variables_list_therapist}, Appendix). The resulting list comprises 15 data categories spanning: (1) patient physiological discomfort, (2) patient–avatar communication dynamics, and (3) procedural and environmental session parameters.

A key guiding principle throughout was to avoid interpretation, deduction, or diagnosis, instead maintaining descriptive, objective, and neutral observations. For example: \textit{``In a 45-minute session, he completed only 6 activities, compared to 10 activities in each of the previous 12 sessions of equal duration.''}

\subsection{Analysis of Real Clinical Notes}
\label{subsubsec:analysis_real_notes}

To better characterize speech therapists' reporting needs for remote remediation, we analyzed a small corpus of clinical notes provided by our two collaborating therapists, comprising:

\begin{itemize}[noitemsep,topsep=2pt]
\item Eight speech therapy assessment report conclusions (including diagnostic conclusions and therapeutic plans), plus one inter-colleague transmission note.
\item Clinical observations gathered by a collaborating speech therapist during a study comparing (1) in-office cognitive training exclusively with the therapist and (2) mixed training combining in-office and autonomous home practice, drawing on feedback from four speech therapists working with MCI and Alzheimer's patients.
\end{itemize}

We analyzed this corpus in a working session with the collaborating therapists to identify domain-specific vocabulary and essential reporting elements. Through iterative review, we extracted frequently occurring keywords and observation elements, then established precise definitions using therapist-validated examples. The resulting taxonomy (Table~\ref{tab:theradia_note_vocabulary}, Appendix) organizes vocabulary into \textbf{ten categories}: \textit{comprehension}, \textit{production}, \textit{emotion}, \textit{execution}, \textit{attention}, \textit{behavior}, \textit{motivation}, \textit{memory}, \textit{reasoning}, and \textit{self-evaluation}. Certain concepts in this framework (e.g., praxis disorders) cannot be reliably inferred from behavioral data alone in remote observation contexts.

\subsection{Feature Selection and Taxonomy}
\label{subsec:theradia_taxonomy}

Synthesizing (1) the remote clinical observation variables from Section~\ref{subsubsec:remote_variables} and (2) the domain vocabulary from Section~\ref{subsubsec:analysis_real_notes}, we developed a reporting framework comprising 11 categories:

\begin{itemize}[noitemsep,topsep=2pt]
    \item \textbf{Comprehension}: The patient's understanding of dialogue, words, and exercise instructions, including instances where the patient explicitly signals confusion.
    \item \textbf{Production}: Lexical, semantic, syntactic, and phonological aspects of patient speech, including informativeness and use of periphrases.
    \item \textbf{Communication}: Patient-initiated communication and responses to the virtual assistant, including response latency.
    \item \textbf{Emotion}: The patient's psycho-affective states throughout the session.
    \item \textbf{Execution}: Patient involvement, contextual factors, and overall exercise performance.
    \item \textbf{Attention}: The patient's concentration on tasks and exercises.
    \item \textbf{Behavior}: Training-specific behavior and any discomfort experienced.
    \item \textbf{Motivation}: Willingness to perform exercises and engage with their objectives.
    \item \textbf{Memory}: Mnesic or gnosic difficulties evidenced during the session (e.g., forgetting).
    \item \textbf{Reasoning}: Quality of patient remarks and capacity to reflect and organize thoughts, as reflected in interaction.
    \item \textbf{Self-evaluation}: The patient's perception of their own performance and affective state.
\end{itemize}

Table~\ref{tab:taxonomy_summary} summarizes these categories, the elements they cover, and their feasibility for automated reporting (detailed definitions and source provenance are provided in Table~\ref{tab:theradia_taxonomy}, Appendix). Two types of constraints limit automated reportability:

\begin{table*}[ht]
  \centering
  \footnotesize
    \caption{Summary of the 11-category taxonomy: key elements, automated reportability ($\checkmark$ = fully automatable; $\sim$ = partial; $\times$ = not automatable), and corresponding sections in the generated report.}
  \begin{tabular}{p{2.3cm} p{5.2cm} c p{2.8cm}}
\toprule
\textbf{Category} & \textbf{Key elements} & \textbf{Auto.} & \textbf{Report section} \\
\midrule
Comprehension      & Word/phrase understanding, question responses               & $\times$ & --- \\
Production         & Lexicon, syntax, informativeness, phonological errors       & $\sim$   & \textit{Language}$^\dagger$ \\
Communication      & Patient-initiated exchanges, response timing                & $\times$ & --- \\
Emotion            & Psycho-affective state                                     & \checkmark & \textit{Affect} \\
Execution          & Session date/duration, activity counts, success rate        & $\sim$   & \textit{Contextual info.} + \textit{Results}$^\ddagger$ \\
Attention          & Focus stability, selective/divided attention                & $\times$ & --- \\
Behavior           & Praxis, initiation, training-specific conduct               & $\times$ & --- \\
Motivation         & Willingness, eagerness to understand                        & $\times$ & --- \\
Memory             & Mnesic/gnosic disorders, forgetting                         & $\times$ & --- \\
Reasoning          & Pertinent remarks, logical reflection                       & $\times$ & --- \\
Self-evaluation    & Perceived performance, affective self-rating                & $\times$ & --- \\
\bottomrule
\multicolumn{4}{p{13cm}}{\footnotesize $^\dagger$ Proxies only: lexical diversity (TTR), lexical density, speech rate, utterance length — qualitative aspects (semantics, phonology) excluded.} \\
\multicolumn{4}{p{13cm}}{\footnotesize $^\ddagger$ Contextual and performance sub-elements are automatable; training environment, third-party presence, and network issues are not.} \\
\end{tabular}

  \label{tab:taxonomy_summary}
\end{table*}

\begin{itemize}[noitemsep,topsep=2pt]
    \item \textbf{Data constraints:} Not all key elements are captured in the THERADIA data. For instance, patient attention and concentration cannot be assessed from available outputs.
    \item \textbf{Detection gaps:} A fundamental asymmetry exists between what a human observer can readily infer and what a machine can automatically detect. For example, determining whether a patient is alone or in a shared room, or detecting external interruptions such as visits, phone calls, or pain episodes, remains beyond reliable automatic extraction.
\end{itemize}

Following expert consultation, we identified extractable and predictable features from the THERADIA data and refined the taxonomy through iterative discussions. Two categories were excluded from the final taxonomy:

\begin{itemize}[noitemsep,topsep=2pt]
    \item \textbf{Exclusion of \textit{Communication}:} Patient responses were often excessively long for report inclusion. Additionally, response delay calculations proved unreliable, as timestamps marked only the onset of avatar speech rather than its end; estimates based on average speech rates introduced unacceptable uncertainty.
    \item \textbf{Exclusion of \textit{Comprehension}:} Relevant data could not be extracted reliably from available sources.
\end{itemize}

The final taxonomy comprises four categories, the first two derived from the \textit{Execution} category:

\begin{itemize}[noitemsep,topsep=2pt]
    \item \textbf{Contextual Information:} Date and duration of the session, and the number of exercises performed with their repetition structure.
    \item \textbf{Results:} Number of successful, moderately successful, and failed exercises, and the overall session success rate.
    \item \textbf{Affect:} Emotional states of the participant during the session.
    \item \textbf{Language:} Linguistic features extracted from spoken interaction, comprising: (1) \textit{lexical} characteristic: vocabulary size, lexical diversity, lexical density; (2) \textit{prosodic} characteristics: total speaking time, speech rate; (3) \textit{syntactic} characteristics: mean utterance length and duration.
\end{itemize}

\section{Methods: Expert-Based Clinical Report Generation}
\label{sec:theradia_methods}

\begin{figure*}
    \centering
    \includegraphics[width=.9\textwidth]{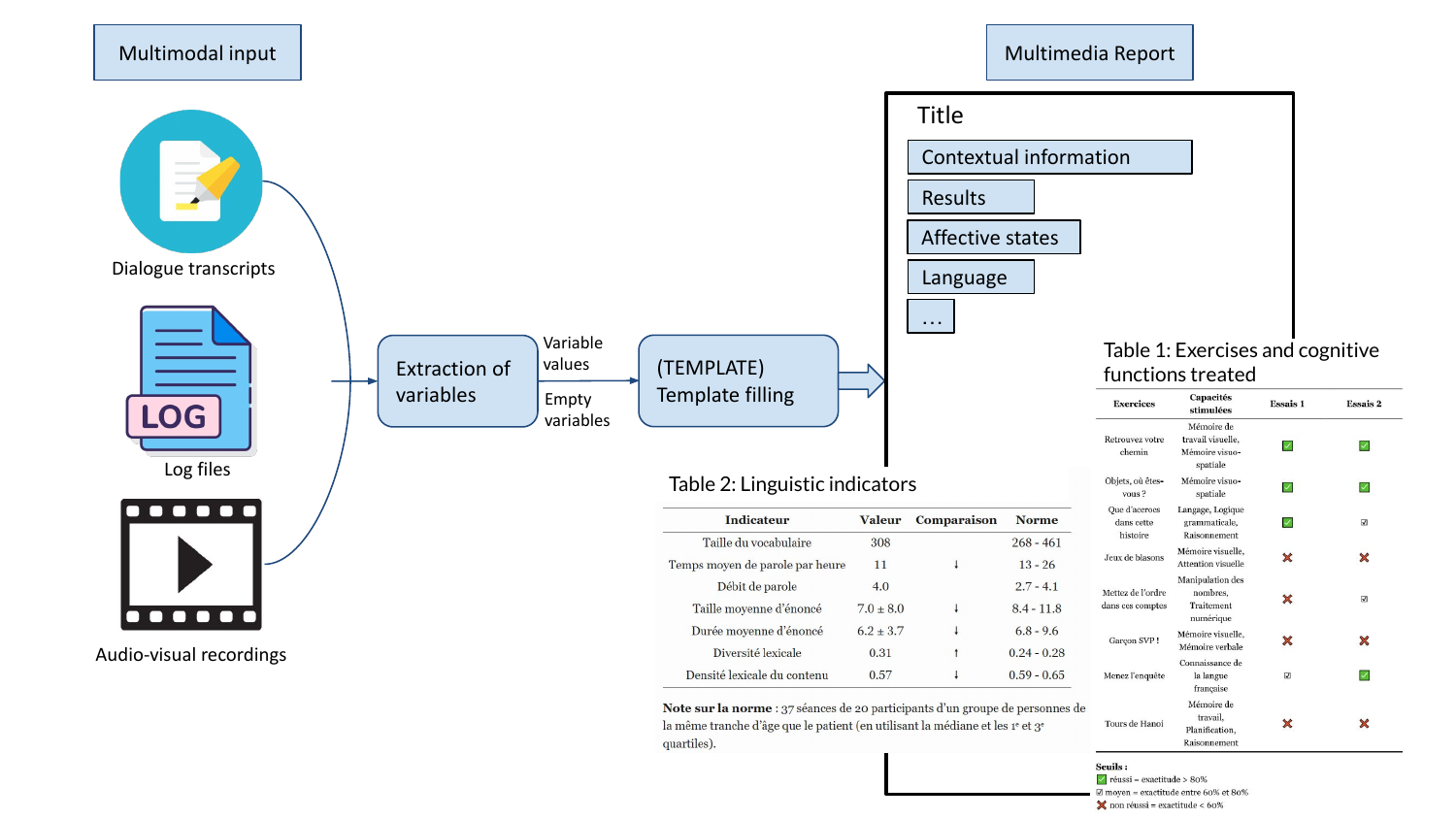}
    \caption{Template-based report generation system. The pipeline integrates multiple data sources (dialogue transcripts, log files, and audio-visual recordings). Key variables, pre-selected by speech therapy experts, are extracted and mapped to predefined template sentences to produce structured clinical reports.}
    \label{fig:theradia_system_generation}
\end{figure*}

For automated remediation session report generation, we developed a template-based system with two key advantages:

\begin{itemize}[noitemsep,topsep=2pt]
    \item \textbf{Explainability}: The transparent template structure supports clear communication with non-expert stakeholders and enables direct integration of clinical expertise.
    \item \textbf{Data efficiency}: The approach eliminates dependency on the large training datasets typically required by machine learning methods.
\end{itemize}

The system generates reports by populating predefined templates with clinically relevant features extracted from session data (dialogue transcripts, log files, and audio-visual recordings), following the taxonomy defined in Section~\ref{sec:theradia_report_form}. As illustrated in Figure~\ref{fig:theradia_system_generation}, the process involves: (1) extracting and quantifying features according to the final taxonomy, and (2) mapping these values to corresponding template sentences. The output report is organized into four taxonomy-aligned sections: \textit{Contextual Information}, \textit{Results}, \textit{Affective States}, and \textit{Language}, supplemented by two tables: one detailing exercises and the cognitive functions they target, and one summarizing linguistic indicators.

\subsection{Iterative Design with Expert Collaborators}

The template-based system was developed through an iterative, expert-guided design process. We began by identifying relevant variables from the THERADIA corpus (dialogue transcripts, log files, and audiovisual recordings), which were used to populate a Markdown template subsequently converted to HTML for presentation.

The system underwent multiple refinement cycles with speech therapists over approximately one year. At each cycle, we presented generated report examples and incorporated therapist feedback into both the template content and the overall report structure. This process yielded concrete insights into the optimal format for virtual-assistant-guided remediation session reports and the system requirements needed to meet clinical standards.

\subsection{Report Structure and Content}

Building on the taxonomy, we designed a structured report comprising four thematic sections (Table~\ref{tab:theradia_descriptors}), two summary tables, and an appendix. Data sources include session logs, dialogue transcripts, and audiovisual recordings, complemented by external \textit{exercise metadata} specifying the cognitive abilities targeted by each exercise.

\begin{table*}[ht]
  \centering
  \footnotesize
  \caption{Variables used in THERADIA report generation, with corresponding template sentences and data sources. Asterisks (*) indicate predicted values. Original templates were developed in French; English translations are provided for accessibility.}
  \begin{tabular}{p{3cm}p{3.4cm}p{8.6cm}}
      \toprule
    \textbf{Source} & \textbf{Variables} & \textbf{Templates} \\
    \hline
\rowcolor{lightgray}
      \multicolumn{3}{c}{\textbf{Paragraph: Contextual Information}}\\
      Log files & date\_session\_string, textual\_start\_time, nb\_activities, nb\_exercises, duration\_session\_str & The session on \{\} took place around \{\}. During this session, the patient completed \{\} activities (\{\} exercises performed twice) over \{\}. Table 1 summarizes the cognitive functions targeted and the results of the activities. \\
      \hline
\rowcolor{lightgray}
      \multicolumn{3}{c}{\textbf{Paragraph: Results}} \\
     Log files & num\_failed & Among these activities: \{\} activities were not successful (correct response rate $<$ 60\%). \\
    Log files, exercise information & num\_partial & \{\} activities were partially successful (correct response rate between 60\% and 80\%). \\
    & & The remaining activities showed completely satisfactory results (correct response rate $>$ 80\%). \\
     \midrule
     Log files & success\_rate & The success rate for the exercises is \{\}\%. \\
      \midrule
     Log files & exo\_failed & The exercises that were not successful are: \{\}. \\
      \hline
\rowcolor{lightgray}
    \multicolumn{3}{c}{\textbf{Paragraph: Affect}}\\
     Audiovisual recordings & emo\_state* & During the session, the patient appeared particularly \{\} (\{\}, but also \{\}) compared to the emotions expressed by patients in the same group. \\
      \hline
\rowcolor{lightgray} 
    \multicolumn{3}{c}{\textbf{Paragraph: Language}}\\
     & & Table 2 below presents the values of the linguistic indicators computed from the patient's utterances during the interaction. Explanations of the different indicators are provided in the Appendix. \\
    Dialogue transcripts & indicator\_higher & Compared to the norm, the value of "\{\}" is higher. \\
    & indicator\_lower & Conversely, the value of "\{\}" is lower. \\
    \bottomrule
\end{tabular}
  \label{tab:theradia_descriptors}
\end{table*}

The \textbf{Contextual Information} section covers the date, time, and duration of the session, and the number of activities and exercises performed. It also introduces the first table, listing exercise names and the cognitive functions they address. Relevant variables include: \textit{date\_session\_string}, \textit{textual\_start\_time}, \textit{nb\_activities,}, \textit{nb\_exercises}, and \textit{duration\_session\_str}.

The \textbf{Results} section reports: (1) separate counts of failed (\textit{num\_failed}) and moderately successful (\textit{num\_partial}) activities; (2) the overall session success rate (\textit{success\_rate}); and (3) a list of failed exercises with their names and order in the session (\textit{exo\_failed}), e.g., \textit{Exo 3 Que\_d'accros ($6^{e}$ activité)}.

The \textbf{Affect} section describes the salient emotions expressed by the participant during the session (\textit{emo\_state}), with positive and negative emotions reported separately where both are present. The affect analysis methods are described in Section~\ref{subsec:theradia_methods_emotion_aware}.

The \textbf{Language} section presents a linguistic analysis organized around the second summary table, reporting variables that fall above (\textit{indicator\_higher}) or below (\textit{indicator\_lower}) population norms. A detailed account of the linguistic analysis is provided in Appendix~\ref{appendix:theradia_linguistic}.

The two summary tables serve distinct functions. \textit{Table~1: Exercises and Targeted Cognitive Functions} provides a structured overview of completed exercises, repetition counts, performance results, and corresponding cognitive domains. \textit{Table~2: Linguistic Indicators} presents quantitative language analysis results complementing the \textit{Language} section. An appendix defining all linguistic metrics and their clinical significance is included at the end of the report.

\subsection{Extracted Features}

As shown in Table~\ref{tab:theradia_descriptors}, the report structure includes variables requiring computational extraction from source files. Below we detail the extraction methods for \textit{Language} and \textit{Affect} variables.

\subsubsection{Extraction of Language-related Variables}

While machines cannot replicate all clinical inferences made by speech therapists, they can compute variables that are either impractical for human calculation or provide complementary quantitative measures. Linguistic and acoustic analyses, for instance, offer rich information for the automatic monitoring of cognitive-linguistic function, and can complement therapist observations to support more comprehensive patient follow-up.

Following \citet{Review-Assessment-Cognitive-Thought-Disorders}, we categorized analyzable features into: (1) \textbf{text-based} (lexical, syntactic, semantic, and pragmatic) and (2) \textbf{acoustic} (prosodic, spectral, and vocal quality). Although acoustic features can provide convergent supplementary evidence, textual features such as lexical diversity and syntactic complexity more directly reflect cognitive-linguistic change. Given the report objectives and current technical limitations in acoustic analysis, we prioritized text-based features across three dimensions:
\textbf{Lexical diversity} was assessed via \textit{type-to-token ratio} (TTR). \textbf{Lexical density} was measured using two metrics: content density, defined as the ratio of content words (verbs, nouns, adjectives, and adverbs) to total words, and propositional density, defined as the ratio of verbs, adjectives, adverbs, prepositions, and conjunctions to total words; both were computed using part-of-speech (POS) tagging \citep{crabbe-candito-2008-experiences}. \textbf{Syntactic complexity} was approximated by mean utterance length, a computationally tractable alternative to parse-tree-based analyses.

Through iterative expert consultation, we refined these measures into seven indicators reported in \textit{Table~2: Linguistic Indicators}, derived from patient dialogue transcripts:

\begin{enumerate}[noitemsep,topsep=2pt]
    \item \textbf{Vocabulary size:} Number of unique words used.
    \item \textbf{Average speaking time per hour:} Total patient speaking time, normalized to minutes per hour.
    \item \textbf{Speech rate:} Number of phonemes per second.
    \item \textbf{Average utterance length:} Mean number of words per utterance.
    \item \textbf{Average utterance duration:} Mean duration of utterances, in seconds.
    \item \textbf{Lexical diversity:} Ratio of unique words to total words (TTR).
    \item \textbf{Content lexical density:} Ratio of content words (verbs, nouns, adjectives, adverbs) to total words.
\end{enumerate}

These indicators were positively received by therapists during prototype testing and were subsequently integrated into the system. Technical implementation details are provided in Appendix~\ref{appendix:theradia_linguistic}.

\subsubsection{Extraction of Affect-related Variables}
\label{subsec:theradia_methods_emotion_aware}

Affective content can offer meaningful insights for analyzing, monitoring, and supporting human interactions \citep{zhou-etal-2024-psentscore-evaluating, zhou:tel-05101086}. In particular, a patient's emotional states can help speech therapists plan follow-up actions. As shown in Table~\ref{tab:theradia_descriptors}, the report includes a paragraph describing the patient's salient emotions using the template: \textit{``During the session, the patient was particularly {} ({}, but also {}) compared to the emotions felt by other patients in the same group.''} To populate this template, we apply emotion recognition models to detect and categorize the patient's emotional states throughout the session.

\paragraph{Multimodal Emotion Recognition Model}

We implemented a multimodal (text--audio--video) emotion recognition model developed within the THERADIA project \citep{Fournier_et_al_2025}, trained and evaluated on the THERADIA-WoZ corpus. This corpus contains 39.5 hours of annotated multimodal interactions from two groups of senior participants (52 neurotypical, 9 with MCI) performing Wizard-of-Oz-assisted computerized cognitive training (CCT) exercises guided by a tele-operated virtual assistant.

The model recognizes ten affective labels relevant to AI-assisted CCT: five positive (\textit{relaxed}, \textit{interested}, \textit{satisfied}, \textit{confident}, \textit{happy}) and five negative (\textit{frustrated}, \textit{surprised}, \textit{annoyed}, \textit{desperate}, \textit{anxious}). It processes sequential inputs through modality-specific feature extractors:

\begin{itemize}[noitemsep,topsep=2pt]
    \item \textbf{Text:} A BERT model \citep{devlin-etal-2019-bert} fine-tuned for sentiment analysis across six languages (English, Dutch, German, French, Spanish, and Italian).\footnote{\url{https://huggingface.co/nlptown/bert-base-multilingual-uncased-sentiment}}
    \item \textbf{Audio:} Wav2Vec2 \citep{10.5555/3495724.3496768} pre-trained on 56 languages including French, and fine-tuned for speech recognition, used here as a feature extractor for acoustic-affective representations.\footnote{\url{https://huggingface.co/voidful/wav2vec2-xlsr-multilingual-56}}
    \item \textbf{Video:} CLIP \citep{pmlr-v139-radford21a}, a vision–-language model trained with contrastive learning, used here to extract frame-level visual features for facial expression analysis.\footnote{\url{https://openai.com/research/clip}}
\end{itemize}

Modality-specific representations are combined through a Multi-Layer Perceptron (MLP) with a hidden layer sized at half the input dimension. The model outputs both probability scores and binary predictions (via learned per-label thresholds) for each affective label.

\paragraph{Model Performance}
The multimodal system achieves a mean Concordance Correlation Coefficient (CCC) of 0.380 for intensity regression and a mean Unweighted Average Recall (UAR) of 71.6\% for presence detection (derived from thresholded regression) across the ten affective labels \citep{Fournier_et_al_2025}, outperforming unimodal baselines (text: CCC 0.229 / UAR 67.7\%; audio: CCC 0.280 / UAR 67.3\%; video: CCC 0.255 / UAR 62.7\%). Per-label classification UAR ranges from 62.3\% (\textit{surprised}) to 76.7\% (\textit{happy}), with all labels exceeding chance level. These results were obtained on a held-out test set from the THERADIA-WoZ corpus under the same experimental conditions as the training data.

We acknowledge that these performance figures reflect in-domain evaluation on the THERADIA-WoZ corpus, and that no external validation on independent clinical data has been conducted. Affect recognition therefore represents the component of the pipeline most subject to uncertainty, and its outputs should be interpreted by clinicians as indicative rather than definitive. To limit the propagation of low-confidence predictions into the report, only emotions whose session-level intensity is statistically significantly elevated relative to population norms (Bonferroni-corrected Z-test, see below) are reported, providing an implicit confidence threshold. Clinician review of the affect paragraph is strongly recommended as part of any deployment workflow.

\paragraph{Determining Salient Emotions for Reporting}

To populate the affect template, we identify emotionally salient states by comparing patient affect against population norms derived from the THERADIA-WoZ test set (13 subjects, 17 sessions), predicted by the same multimodal model. 
This population-level comparison was adopted as the reference standard because the current dataset consists of single sessions per MCI participant, making intra-patient longitudinal baselines unavailable. 

Significance is assessed using a right-tailed Z-test (\(Z = \frac{\bar{X} - \mu}{\sigma / \sqrt{n}}\)), where \(\bar{X}\) is the session mean intensity, \(\mu\) and \(\sigma\) are the population mean and standard deviation, and \(n\) is the number of sequences. Sessions typically contain more than 100 sequences, satisfying the normality assumption. 
For each target session, pairwise comparisons are performed between each emotion's distribution in that session and the corresponding distributions across all other subjects in the test set. To control for false positives arising from multiple simultaneous comparisons, \textit{p}-values are corrected using the Bonferroni correction.

\section{Using LLMs for Clinical Report Generation}
\label{sec:theradia_llm}

Due to the absence of reference reports, fine-tuning a language model for this task is not feasible; we therefore rely on zero-shot prompting. To mitigate hallucination risk, generation is grounded in pre-extracted, expert-validated structured variables rather than raw session transcripts, ensuring the model can only report facts already verified by the rule-based pipeline and enabling a fair factual comparison with template-based outputs. We conducted experiments using GPT-4\footnote{Experiments were conducted in 2024 using the then-state-of-the-art \textsc{GPT-4}; see the Limitations section for a discussion of this work as a controlled technological snapshot.}, storing the variables listed in Table~\ref{tab:theradia_descriptors} and their values in JSON format as structured input to the model.

\subsection{Model and Parameters}

Experiments were conducted using the OpenAI \textsc{GPT-4} API\footnote{\url{https://platform.openai.com/docs/models/gpt-4}} (version \textsc{gpt-4-0613}), one of the most capable publicly available models at the time of the study (late 2023). The model supports a maximum context length of 8,192 tokens. To ensure deterministic outputs, the temperature parameter was set to $0$; all other parameters were left at their default values.

\subsection{Prompt Design}

We evaluated three prompting strategies during preliminary experiments with ChatGPT:

\begin{itemize}[noitemsep,topsep=2pt]
    \item \textbf{Guideline Prompt:} Report-writing guidelines (including key criteria) provided alongside raw session transcripts and log files.
    \item \textbf{Augmented Guideline Prompt:} Guidelines combined with pre-extracted variable values.
    \item \textbf{Variables-specified Prompt:} Structured variables with explanatory notes provided alongside their pre-extracted values.
\end{itemize}

We selected the \textit{Variables-specified Prompt} with pre-extracted values for three reasons. First, processing raw transcripts and log files directly leads to excessive token lengths that frequently exceed GPT-4's context window; using curated variables reduces both computational cost and input complexity. Second, using identical variables to those in the template-based system ensures methodological consistency across generation approaches. Third, pre-selected variables provide standardized, reliable input that encodes the domain expertise of the variable selection process and aligns with clinical best practices: prioritizing objective, descriptive reporting while avoiding diagnostic inference or subjective interpretation.

The variables are provided in JSON format, and the prompt is shown in Table~\ref{tab:prompt_theradia_rg}\footnote{Token count may vary depending on the content of the \textit{Variables} field; total input tokens including prompt text are approximately 1,600.}. The prompt opens with a brief task description, followed by variable definitions and the values populating the two tables. The model is explicitly instructed to produce a factual report free of diagnostic interpretation, formatted in Markdown, consistent with the template-based system, to support section and subsection headings.

\begin{table}[!ht]
\centering
\caption{Meta prompt used to generate reports with GPT-4 (translated from French). ``\&\#129047;'' is the HTML entity for the Unicode character “↓”, and ``\&\#129045;'' is that for “↑”.}
\scriptsize
\begin{tcolorbox}[
  colback=steel!5,
  colframe=steel!75,
  title=Prompt for Cognitive Remediation Session Report Generation,
  fonttitle=\bfseries,
  ]

You must summarize a cognitive remediation session for a speech-language pathologist. A cognitive remediation session involves a patient and a virtual assistant. This assistant offers exercises to the patient to stimulate various cognitive functions such as language, planning, and memory.

To summarize the session, you will use the information contained in the JSON file below and write a report. The file presents the different variables with explanations in parentheses:

“date\_session\_string” (Session date) \\
“textual\_start\_time” (Session start time) \\
“nb\_activities” (Number of activities completed) \\
“nb\_exercises” (Number of exercises completed) \\
“duration\_session\_str” (Session duration) \\
“num\_failed” (Number of failed activities) \\
“num\_partial” (Number of partially successful activities) \\
“success\_rate” (Success rate - successful activities / total activities) \\
“exo\_failed” (Failed activities) \\
“salientEmotions” (Emotions particularly expressed by the patient compared to patients in the same group)

“Exo\_results\_TableDict” (Exercises and cognitive functions addressed) - Can be used to generate a table \\
(list\_of\_strings = ["Exercise", "Cognitive skills stimulated", "Attempt 1", "Attempt 2"], elements are "Exercise names", "Cognitive skills stimulated by each exercise", "N1 (Attempt 1): Results of the first attempt (activity) of an exercise", and "N2 (Attempt 2): Results of the second attempt of an exercise".) \\
(Thresholds: "{\color{green}\checkmark} successful = accuracy > 80\%", "{\color{orange}\checkmark} partial = accuracy between 60\% and 80\%", "{\color{red}\checkmark} unsuccessful = accuracy < 60\%") \\

“TableDict” (Linguistic indicators) - Can be used to generate a table \\
(list\_of\_strings = ["Indicator", "Value", "Comparison", "Norm"], elements are "linguistic indicators", "values of these indicators", "a comparison of the value to the norm (above or below)", "the norm for this indicator".) \\
(comparison = "\&\#129047;" means “↓”, the value is below the norm; comparison = "\&\#129045;" means “↑”, the value is above the norm) \\
(Note on the norm: 37 sessions from 20 participants in a group of individuals of the same age range as the patient (using median and 1\textsuperscript{st} and 3\textsuperscript{rd} quartiles).)

(Explanations of linguistic indicators: 
    Vocabulary size: number of unique words. 
    Mean speech time per hour: in minutes. 
    Speech rate: number of phonemes per unit of time (seconds). 
    Mean utterance length: average number of words per utterance. 
    Mean utterance duration: in seconds. 
    Lexical diversity: number of unique words divided by total number of words. 
    Content lexical density: number of content words (verbs, nouns, adjectives, adverbs) divided by total number of words.)

The report must not contain any diagnosis or interpretation but should focus on factual data. It is better to remain descriptive, objective, and neutral.

Example:

JSON:

\{\{Variables\}\}

Please write a report to inform a speech-language pathologist about the course of this session. Provide your response in a Markdown code block.

\end{tcolorbox}
\label{tab:prompt_theradia_rg}
\end{table}

\section{Generated Reports and Evaluation}
\label{sec:theradia_evaluation}

Examples of template-based and GPT-4 reports are presented in Sections~\ref{subsec:theradia_report_generated} and~\ref{subsec:theradia_generated_llms}, respectively. The human evaluation design and results follow in Sections~\ref{subsec:human_evaluation} and~\ref{subsec:evaluation_results}.

\subsection{Examples of Template-based Reports}
\label{subsec:theradia_report_generated}

\begin{figure*}[!ht]
    \centering
    \includegraphics[width=.9\textwidth]{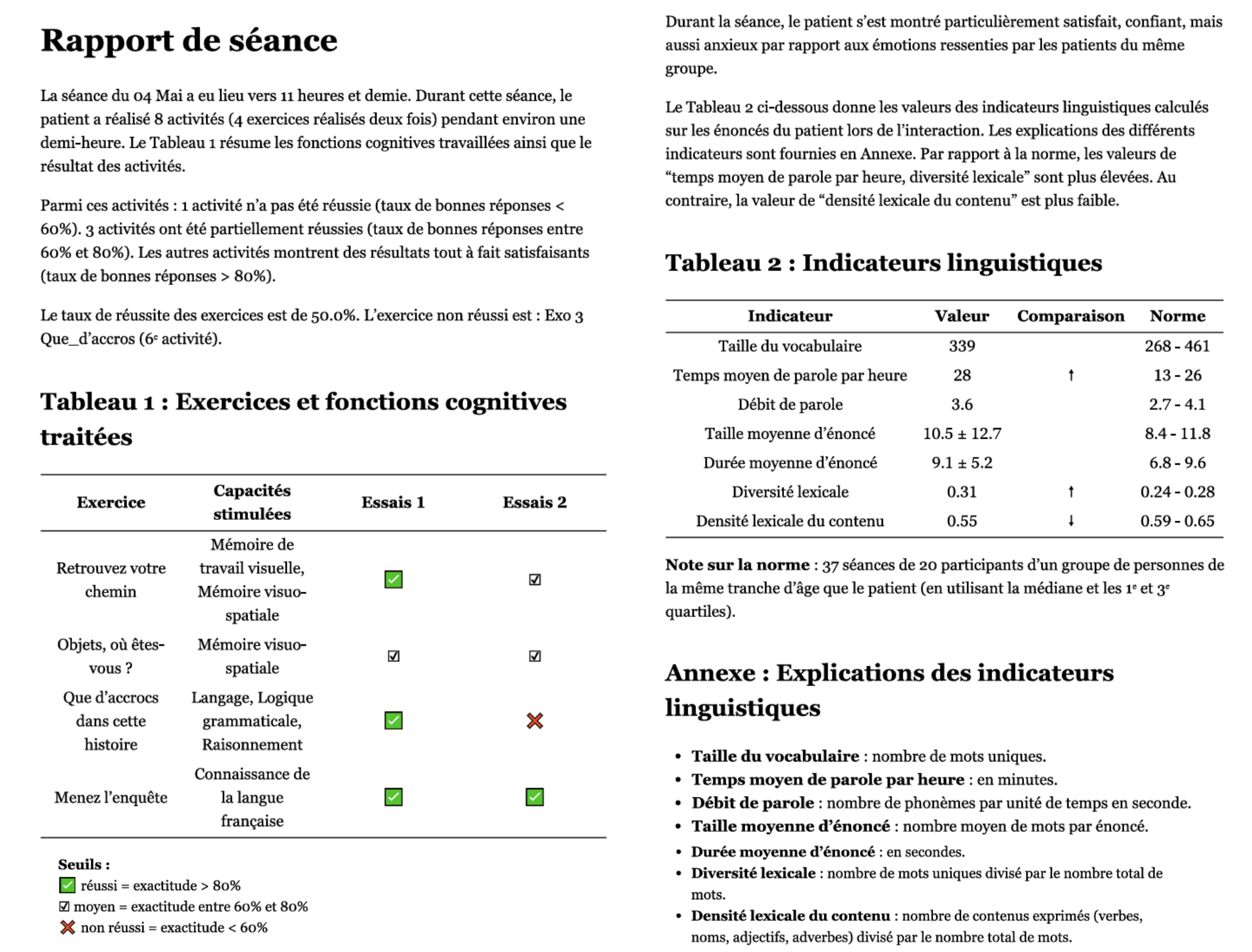}
    \caption{Example of report generated by the template-based system for participant \textit{M01E} (first session).}
    \label{fig:report_example_template}
\end{figure*}

Figure~\ref{fig:report_example_template} shows a representative report for MCI participant \textit{M01E}. Each report comprises three components: (1) a structured textual summary organized into \textit{Contextual Information}, \textit{Results}, \textit{Affect}, and \textit{Language}, populated with session-derived variables (Table \ref{tab:theradia_descriptors}); (2) two supporting tables covering exercise results and linguistic indicators with normative values; and (3) an appendix explaining specialized linguistic terminology for therapists unfamiliar with these metrics. All template-generated reports share this structure; content varies according to the session data.

\subsection{Examples of LLM-Generated Reports}
\label{subsec:theradia_generated_llms}

\begin{figure*}[!ht]
    \centering
    \includegraphics[width=.45\textwidth]{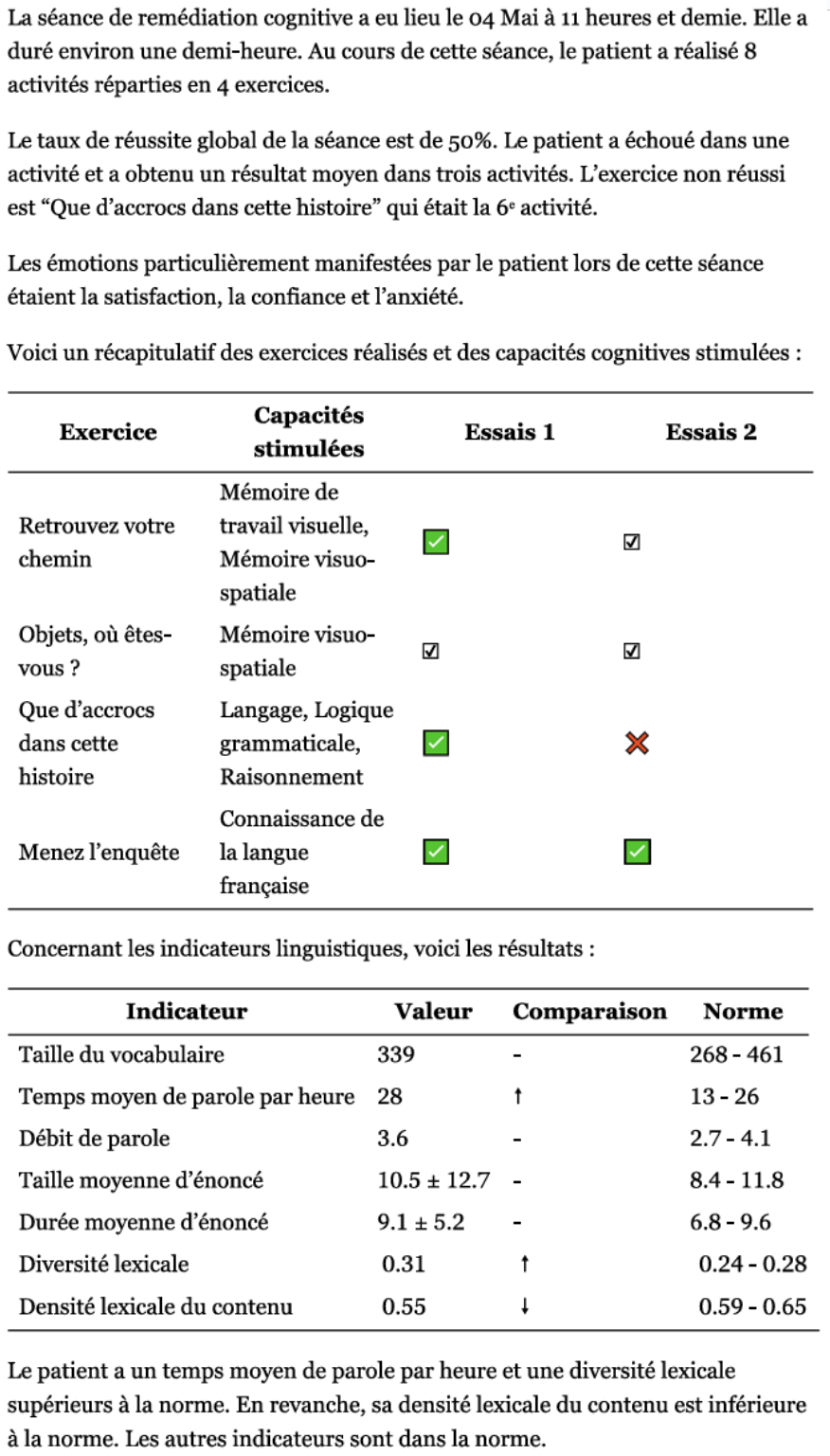}
    \caption{Example of report generated by GPT-4 for participant \textit{M01E} (first session).}
    \label{fig:report_example_llm}
\end{figure*}

Figure~\ref{fig:report_example_llm} shows the GPT-4 report for the same participant. It is notably shorter than its template-based counterpart: table captions and the linguistic indicator appendix are absent despite being specified in the prompt, a consistent pattern across GPT-4 outputs. In one case, a report contained text only with no tables, likely because nearly all exercise results were successful, illustrating the model's ability to adapt its output structure to the input data.

\subsection{Human Evaluation by Questionnaire}
\label{subsec:human_evaluation}

Automatic evaluation of clinical reports is critical for ensuring their validity and reliability \citep{zhou-etal-2023-survey}. Given the absence of reference reports and expert consensus, we adapted the evaluation framework of \citet{savkov-etal-2022-consultation}, using targeted questions to assess report quality across five domains defined with our collaborating speech therapists: (1) general aspects, (2) execution, (3) affective states, (4) performance and difficulties, and (5) language production. All reports and questionnaires were in French (the evaluators' native language); English translations are provided below.

The questionnaire comprises \textbf{nine core metrics} rated on a 1–5 Likert scale: (1) \textit{Fluidity}, (2) \textit{Conciseness}, (3) \textit{Relevance}, (4) \textit{Coherence}, (5) \textit{Session Information}, (6) \textit{Affective States}, (7) \textit{Results}, (8) \textit{Cognitive Functions}, and (9) \textit{Linguistic Indicators}, plus one open-ended comment field. Questions were selected to cover essential aspects, with each item addressing content produced by at least one system, and the total kept to around ten. The scale presented the highest rating (5, best outcome) first and the lowest (1, worst) last.

\begin{enumerate}[noitemsep,topsep=2pt]
    \item \textbf{Fluidity:} Le rapport vous paraît-il fluide ? [Does the report appear fluid to you?]
    \item \textbf{Conciseness:} Le rapport est-il concis ? [Is the report concise?] 
    \item \textbf{Relevance:} Les informations données dans le rapport sont-elles pertinentes ? [Is the information provided in the report relevant?]
    \item \textbf{Coherence:} Le rapport vous parait-il cohérent ? [Does the report appear coherent to you?] 
    \item \textbf{Session Info:} Le rapport fournit-il des informations pertinentes sur le déroulement de la séance ? Indiquez toute information qui vous semblerait pertinente à ajouter. [Does the report provide relevant information about the progress of the session? Indicate any information you feel should be added.] 
    \item \textbf{Affective States:} Est-ce que le rapport semble bien décrire les états affectifs exprimés par le participant au cours de la séance ? [Does the report seem to describe well the affective states expressed by the participant during the session?] 
    \item \textbf{Results:} Est-ce que les résultats des exercices sont présentés de façon claire ? [Are the results of the exercises presented clearly?] 
    \item \textbf{Cognitive Functions:} Le rapport précise-t-il les fonctions cognitives et permet-il d'évaluer les fonctions cognitives du patient pendant la séance ? [Does the report specify the cognitive functions and allow for the evaluation of the patient's cognitive functions during the session?] 
    \item \textbf{Linguistic Indicators:} Les indicateurs linguistiques présentés dans le rapport sont-ils utiles ?  [Are the linguistic indicators presented in the report useful?] 
    \item Est-ce que vous avez des commentaires à ajouter ? [Do you have any comments to add?] 
\end{enumerate}

We generated reports for five MCI participants (M01E – M05E) using both approaches (template-based and GPT-4) yielding ten reports for comparative assessment, administered via \textit{LimeSurvey}. Evaluators accessed each report alongside the corresponding questions through a single link, and the scale was consistently presented with the highest rating first to facilitate cognitive processing.

The evaluation task required experts to review five reports and answer ten questions per report. Each questionnaire page presented one generated report alongside its corresponding questions. Eight evaluators participated: four speech therapists and four final-year speech therapy students. We created two survey versions with alternating report types (template-based and GPT-4 generated), always starting with a template-based example: version one contained three template-based and two GPT-4 reports; version two contained two template-based and three GPT-4 reports. Four evaluators completed each version. Participants received a 30-euro gift voucher upon completion. The recruited speech therapists had diverse work experience; notably, one had prior experience using \textit{HappyNeuron} (the cognitive training software used to collect the THERADIA corpus data) with her patients.

\subsection{Evaluation Results}
\label{subsec:evaluation_results}

\subsubsection{Quantitative Results}

\begin{table*}[ht]
\footnotesize
\caption{Human evaluation results of GPT-4 and template-based system on various criteria (mean $\pm$ std, scale 1--5). $^\ast$p$<$0.05, Mann-Whitney U test (two-sided, uncorrected); no dimension reached significance after Bonferroni correction. \textit{Session} stands for \textit{Session Information}, \textit{Affect} stands for \textit{Affective States}, \textit{CF} stands for \textit{Cognitive Functions}, \textit{LI} stands for \textit{Linguistic Indicators}. The first category presents the overall evaluation results of eight evaluators, including four speech therapists and four late-stage speech therapy students. The last two categories present the evaluation results of speech therapists and students, respectively.}
\resizebox{\textwidth}{!}{\begin{tabular*}{\textheight}{@{\extracolsep\fill}lcccccccccc}
\toprule
& \textbf{Fluidity} & \textbf{Conciseness} & \textbf{Relevance} & \textbf{Coherence} & \textbf{Session} & \textbf{Affect} & \textbf{Results} & \textbf{CF} & \textbf{LI} & \textbf{Overall} \\ 
\midrule
\rowcolor{lightgray}
\multicolumn{11}{|l|}{\textbf{All} (eight evaluators)} \\ 
\midrule
\textbf{Template} & \textbf{4.50$\pm$0.61}$^\ast$ & 4.15$\pm$1.18 & 3.85$\pm$1.04 & \textbf{4.25$\pm$0.72} & 3.65$\pm$0.81 & 3.60$\pm$0.94 & \textbf{4.45$\pm$0.94} & 3.50$\pm$1.19 & 3.25$\pm$1.52 & 3.91$\pm$0.65 \\ 
\midrule
\textbf{GPT4} & 3.65$\pm$1.27 & \textbf{4.70$\pm$0.47} & 3.90$\pm$0.97 & 3.85$\pm$0.88 & 3.45$\pm$1.0 & 3.70$\pm$1.13 & 3.70$\pm$1.30 & 3.45$\pm$1.0 & \textbf{3.85$\pm$1.27} & 3.81$\pm$0.75 \\ \midrule
\rowcolor{lightgray}
\multicolumn{11}{|l|}{\textbf{Speech therapists} (four evaluators, two participated in survey n°1 and two participated in survey n°2)} \\ 
\midrule
\textbf{Template} & 4.50$\pm$0.53$^\ast$ & 4.80$\pm$0.42 & 4.20$\pm$1.14 & 4.4$\pm$0.70 & 3.80$\pm$1.03 & 3.9$\pm$0.99 & 4.20$\pm$1.23 & 3.60$\pm$1.26 & 3.40$\pm$1.71 & 4.09$\pm$0.75 \\ 
\midrule
\textbf{GPT4} & 3.20$\pm$1.23 & 4.70$\pm$0.48 & 3.80$\pm$1.23 & 3.50$\pm$1.08 & 3.30$\pm$1.06 & 3.70$\pm$1.25 & 3.50$\pm$1.35 & 3.50$\pm$0.85 & 3.60$\pm$1.51 & 3.64$\pm$0.93 \\ 
\midrule
\rowcolor{lightgray}
\multicolumn{11}{|l|}{\textbf{Students} (four evaluators, two participated in survey n°1 and two participated in survey n°2)} \\ 
\midrule
\textbf{Template} & \textbf{4.50}$\pm$0.71 & 3.50$\pm$1.35 & 3.50$\pm$0.85 & 4.1$\pm$0.74 & 3.50$\pm$0.53 & 3.30$\pm$0.82 & 4.70$\pm$0.48 & 3.40$\pm$1.17 & 3.10$\pm$1.37 & 3.73$\pm$0.52 \\ \midrule
\textbf{GPT4} & 4.10$\pm$1.2 & 4.70$\pm$0.48$^\ast$ & 4.00$\pm$0.67 & 4.20$\pm$0.42 & 3.60$\pm$0.97 & 3.70$\pm$1.06 & 3.90$\pm$1.29 & 3.40$\pm$1.17 & 4.10$\pm$0.99 & 3.97$\pm$0.50 \\ 
\bottomrule
\end{tabular*}
}
\label{tab:theradia_human_evaluation_results}
\end{table*}

Table~\ref{tab:theradia_human_evaluation_results} presents human evaluation scores across the nine criteria, organized by overall results (all eight evaluators) and broken down by speech therapists and students separately. Mann-Whitney U tests revealed no significant differences between systems after Bonferroni correction, though \textit{Fluidity} reached nominal significance (p$<$0.05, uncorrected).

Overall, the template-based system ranks higher on \textit{Fluidity}, \textit{Coherence}, and \textit{Results}; GPT-4 ranks higher on \textit{Conciseness}. A per-criterion reading reveals patterns that reflect the structural properties of each system.

\textit{Fluidity} shows the largest template advantage (4.50 vs.\ 3.65): fixed headings and predictable structure produce a smooth reading experience, whereas GPT-4 reports occasionally show uneven transitions or merged sections. \textit{Coherence} follows similarly (4.25 vs.\ 3.85), especially among speech therapists (4.40 vs.\ 3.50), for whom the logical section sequence provides a familiar and trustworthy structure. On \textit{Results} (4.45 vs.\ 3.70), the template's explicit performance thresholds (successful $>80\%$, partial $60$–$80\%$, failed $<60\%$) give evaluators clear interpretive context — a structural element GPT-4 consistently omitted despite being specified in the prompt, and the single most cited source of dissatisfaction in qualitative feedback. \textit{Relevance} and \textit{Cognitive Functions} are near-tied across both systems (3.85 vs.\ 3.90 and 3.50 vs.\ 3.45), as both draw on the same pre-extracted variables. \textit{Conciseness} favors GPT-4 overall (4.70 vs.\ 4.15), though the gap narrows among speech therapists (4.70 vs.\ 4.80), suggesting experienced clinicians value the template's thoroughness rather than perceiving it as verbosity.

The two most novel dimensions, \textit{Affective States} and \textit{Linguistic Indicators}, yield the most nuanced results. Both systems receive moderate \textit{Affective States} scores (3.60 and 3.70) with the highest standard deviations in the table, indicating strong evaluator disagreement independent of generation approach. Three evaluators found comparison to population norms confusing, arguing that intra-patient longitudinal comparison is more clinically meaningful — pointing to a limitation of the underlying representation rather than generation. For \textit{Linguistic Indicators}, GPT-4 scores higher (3.85 vs.\ 3.25), yet both systems receive the lowest absolute scores overall, reflecting divided expert opinion: evaluators focused on cognitive training questioned the clinical relevance of quantitative lexical metrics for visuospatial or numerical tasks, while those with a stronger linguistic background found them valuable for longitudinal monitoring. This suggests linguistic indicators should be an optional, configurable module rather than a fixed report section.

The small scale of this evaluation (ten reports, eight evaluators) reflects a fundamental constraint of clinical NLG, where structured assessments from certified clinicians require significant coordination effort.
This is consistent with prior work: \citet{PORTET2009789} evaluated BabyTalk with three neonatal nurses, and \citet{savkov-etal-2022-consultation} used a similarly small expert panel. The consistency of directional patterns across all nine dimensions, accompanied by qualitative feedback, suggests the observed differences are meaningful and warrants larger-scale follow-up.

Speech therapists showed a general preference for the template-based system, while students preferred GPT-4, likely reflecting experienced clinicians' preference for structured, comprehensive reports over the shorter outputs favored by younger evaluators. Qualitative feedback is analyzed in the following section.

\subsubsection{Qualitative Results}
Beyond Likert scores, evaluators provided written comments after each question and participated in a follow-up exchange after completing the survey. A consistent theme across feedback was the inherent subjectivity of report evaluation, with experts often holding differing perspectives.

\paragraph{Preferences: Template-based vs.\ GPT-4}
Expert preferences fell into three camps: template-based, GPT-4, or no strong preference. Advocates of the template-based system valued its \textbf{precision}, notably the explicit thresholds (e.g., successful $>80\%$), definitions of linguistic indicators, and structured headings and appendices that make session progression easy to follow. GPT-4 supporters appreciated its \textbf{more natural language}, particularly in conveying affective states and results, and its greater brevity.

\paragraph{Recommendations for Report Generation}
Based on expert feedback, we propose the following recommendations for generating home-based remediation session reports:

\begin{itemize}[noitemsep,topsep=2pt]
    \item \textbf{Longitudinal tracking:} Two experts recommended summarizing performance across sessions (weekly, monthly, or over six months) to track treatment progress and flag fluctuations in patient abilities.
    The results in the table could facilitate comparisons between sessions (e.g., 10 sessions), enhancing longitudinal analysis.
    \item \textbf{Affective states:} Four experts found affective state descriptions too brief or vague. Recommendations included: describing affect per exercise (especially failed ones); avoiding sentences that mix too many emotions simultaneously; limiting reported emotions to two positive and two negative; and noting the patient's condition on arrival. One expert also suggested appending response transcripts to help contextualize failures.
    \item \textbf{Exercise presentation:} Experts recommended clearly describing each exercise's objective and modality, including difficulty level, activated indices, number of answer choices, and instructions, to better contextualize patient performance. A brief appendix overview was suggested for clarity.
    \item \textbf{Results presentation:} Suggestions included providing precise session timestamps, locating failed tasks within the session, specifying error types, and noting affect during specific failures. For example, in a word-selection exercise, knowing which word was incorrectly chosen and at what difficulty level would aid interpretation.
    \item \textbf{Patient information:} Recommendations included adding patient demographics (age, disorders), environmental context (location, noise level, presence of others), and concentration fluctuations across tasks. One expert suggested using "Madame/Monsieur" and a patient photo to personalize reports.
    \item \textbf{Visualization:} Two experts preferred tables and graphics over prose for faster comprehension, recommending visual elements wherever possible.
    \item \textbf{Personalization:} Given varying preferences for report length and detail, experts recommended allowing therapists to select which categories to include before generation.
    \item \textbf{Self-comparison:} Three experts found comparisons to group norms confusing, arguing that intra-patient longitudinal comparison is more clinically meaningful. Recommendations included comparing the patient's current affect or performance to their own previous sessions, and providing timestamps or transcripts when emotional intensity peaks.
    \item \textbf{Conciseness:} Experts recommended avoiding repetition of table captions in the main text, using visual elements in place of prose where possible, and structuring reports as a brief core with a comprehensive appendix.
    \item \textbf{Linguistic indicators:} Experts were evenly divided, those experienced with cognitive exercises preferred qualitative feedback (e.g., patient responses post-exercise), while others valued quantitative lexical metrics, reinforcing the case for making this section configurable.
\end{itemize}

\section{Discussion}
\label{sec:discussion_conclusion}
\subsection{RQ1 — Content Selection: Does the Taxonomy Cover Clinical Needs?}
The four-category taxonomy (Contextual Information, Results, Affect, Language), derived through iterative expert collaboration from an initial 11-category framework, covers the observational dimensions speech therapists identify as clinically relevant while remaining constrained to what can be reliably extracted from available data.
However, evaluation revealed a persistent gap between what the taxonomy captures and what clinicians would ideally receive. Experts consistently emphasized the need to understand \textit{why} an exercise failed, including specific error types, incorrect responses, and contextual factors such as patient fatigue, rather than merely \textit{that} it failed. This motivates two concrete directions: (1) enriching exercise output data with error-level detail, and (2) incorporating patient utterance transcripts for failed exercises as a configurable optional element.
The exclusion of \textit{Communication} and \textit{Comprehension} from the final taxonomy reflects a broader principle: a system should only report what it can extract reliably. Overreporting inferred or uncertain observations risks undermining clinician trust, and assessing motivation, concentration, and comprehension from interaction logs alone underlines the need for advances in multimodal behavior analysis.

\subsection{RQ2 — System Design: Template-based vs.\ LLM-based Generation}
The comparison reveals a clear trade-off between clinical reliability and linguistic quality. The template-based system outperformed GPT-4 on dimensions tied to clinical safety — \textit{Fluidity} (4.50 vs.\ 3.65), \textit{Coherence} (4.25 vs.\ 3.85), and \textit{Results} (4.45 vs.\ 3.70) — advantages traceable to explicit performance thresholds, mandatory section headings, and a fixed document schema. GPT-4 scored higher on \textit{Conciseness} (4.70 vs.\ 4.15) and \textit{Linguistic Indicators} (3.85 vs.\ 3.25), reflecting its capacity for natural prose without formulaic repetition.
This safety/fluency trade-off echoes prior clinical NLG research: rule-based systems such as BabyTalk \citep{PORTET2009789} and STOP \citep{REITER200341} reliably outperform neural alternatives on factual accuracy and structural consistency, while producing less natural text. Zero-shot LLMs add a new dimension (no task-specific training data required) but introduce hallucination risk and, as our results confirm, do not reliably follow all prompt constraints.
The practical implication is that these approaches are \textit{complementary}. Template-based generation suits primary clinical documentation where traceability and threshold transparency are paramount, consistent with the preference of experienced speech therapists (4.09 vs.\ 3.64). LLM-based generation may be better suited for secondary summaries or patient-facing communications. A hybrid architecture using the template as a factual backbone with LLM fluency post-editing represents a promising direction.

\subsection{RQ3 — Evaluation: What the Human Assessment Reveals Beyond Scores}
The divergence between speech therapist and student evaluators, with therapists preferring the template system (4.09 vs.\ 3.64) while students preferred GPT-4 (3.97 vs.\ 3.73), demonstrates that perceived report quality is partly a function of professional experience. Experienced clinicians expect completeness, threshold transparency, and structured organization; students, less bound by such conventions, respond more positively to natural register and brevity.
This has a direct methodological implication: evaluator profile must be controlled for and reported, as aggregating across heterogeneous groups risks obscuring meaningful signal. More broadly, the high subjectivity of quality assessment, reflected in large standard deviations (up to $\pm$1.52 for Linguistic Indicators), argues for evaluation frameworks that separate objective dimensions (factual accuracy, completeness) from stylistic ones (fluency, conciseness). 

\subsection{Generalizability and Broader Implications}
The methodology developed here includes expert elicitation, iterative taxonomy refinement, input-controlled generation, and multi-dimensional human evaluation. This approach is transferable to other digital therapy and remote monitoring contexts, such as telestroke rehabilitation, home physiotherapy, or mental health chatbots, though the domain-specific taxonomy requires re-elicitation for each new application.
A more general finding for expert systems design: the primary clinical utility of automated session reports resides in \textit{structured tabular data} rather than narrative text. Evaluators consistently gave higher and more consistent scores to results tables and exercise summaries than to narrative paragraphs, regardless of generation system, suggesting that future systems should prioritize rich tabular and visual presentation, with narrative text as a complementary interpretive layer.

\section{Conclusion}
\label{sec:conclusion}

This paper investigated automated clinical report generation for home-based cognitive remediation sessions guided by a virtual assistant, addressing a real clinical need: providing speech therapists with succinct, faithful summaries from rich multimodal session logs. Operating in a low-resource setting with no reference reports, we developed a domain taxonomy through iterative collaboration with speech therapists, guiding a rule-based template system that produces structured reports across four dimensions: session context, exercise results, affective states, and linguistic indicators. We compared this system against GPT-4 with zero-shot structured prompting.

The central finding is a clear trade-off between clinical reliability and linguistic quality. The template-based system is more reliable, coherent, and clinically precise, while GPT-4 produces more concise and natural output. Both received positive overall evaluations (mean scores 3.91 and 3.81 out of 5), confirming that the expert-validated taxonomy provides a solid foundation regardless of generation method, consistent with the broader clinical NLG literature. The explainability and traceability of the template-based approach are particularly relevant in clinical deployment contexts, where regulatory expectations and clinician trust requirements favor auditable outputs.

Expert feedback also highlighted several directions for refinement. The affect recognition component relies on group-level norms, which evaluators found less informative than intra-patient longitudinal comparison, pointing to the value of session-to-session tracking. Template structure, while valued for its consistency, can produce formulaic output across reports, motivating configurable modules that allow clinicians to tailor content to their needs. These observations, together with broader evaluator feedback, inform the design recommendations we derive for future systems: configurable report modules, longitudinal aggregation across sessions, richer tabular and graphical presentation of results, exercise-level contextualization of affective states, and optional transcripts for failed exercises.

Three directions are most important for future work. First, the evaluation should be replicated with a larger expert panel across multiple CCT platforms to assess generalizability beyond the THERADIA context. Second, the comparison should be repeated with newer LLMs, and LLM-as-judge evaluation explored as an automated complement to costly human assessment, to track how the safety-fluency trade-off evolves as model capabilities improve. Third, and most clinically impactful, the system should be extended to support longitudinal report generation across sessions, moving beyond the single-session scope of the current work and enabling therapists to monitor patient trajectories and detect performance fluctuations over time.

\bibliographystyle{unsrtnat}
\bibliography{references}  

@InProceedings{10.1007/978-3-030-80285-1_55,
author="Tarpin-Bernard, Franck
and Fruitet, Joan
and Vigne, Jean-Philippe
and Constant, Patrick
and Chainay, Hanna
and Koenig, Olivier
and Ringeval, Fabien
and Bouchot, B{\'e}atrice
and Bailly, G{\'e}rard
and Portet, Fran{\c{c}}ois
and Alisamir, Sina
and Zhou, Yongxin
and Serre, Jean
and Delerue, Vincent
and Fournier, Hippolyte
and Berenger, K{\'e}vin
and Zsoldos, Isabella
and Perrotin, Olivier
and Elisei, Fr{\'e}d{\'e}ric
and Lenglet, Martin
and Puaux, Charles
and Pacheco, L{\'e}o
and Fouillen, M{\'e}lodie
and Ghenassia, Didier",
editor="Ayaz, Hasan
and Asgher, Umer
and Paletta, Lucas",
title="THERADIA: Digital Therapies Augmented by Artificial Intelligence",
booktitle="Advances in Neuroergonomics and Cognitive Engineering",
year="2021",
publisher="Springer International Publishing",
address="Cham",
pages="478--485",
isbn="978-3-030-80285-1"
}

@ARTICLE{Fournier_et_al_2025,
  author={Fournier, Hippolyte and Alisamir, Sina and Azzakhnini, Safaa and Zsoldos, Isabella and Trân, Eléonore and Bailly, Gérard and Elisei, Frédéric and Bouchot, Béatrice and Varini, Brice and Constant, Patrick and Fruitet, Joan and Tarpin-Bernard, Franck and Rossato, Solange and Portet, François and Koenig, Olivier and Chainay, Hanna and Ringeval, Fabien},
  journal={IEEE Transactions on Affective Computing}, 
  title={THERADIA WoZ: An Ecological Corpus for Appraisal-Based Affect Research in Healthcare}, 
  year={2025},
  volume={16},
  number={3},
  pages={2233-2244},
  doi={10.1109/TAFFC.2025.3557465}}

@article{YANG2023100007,
title = {The impact of ChatGPT and LLMs on medical imaging stakeholders: Perspectives and use cases},
journal = {Meta-Radiology},
volume = {1},
number = {1},
pages = {100007},
year = {2023},
issn = {2950-1628},
doi = {https://doi.org/10.1016/j.metrad.2023.100007},
url = {https://www.sciencedirect.com/science/article/pii/S2950162823000073},
author = {Jiancheng Yang and Hongwei Bran Li and Donglai Wei}
}

@article {cognitive_smartphone,
	author = {Klimova, Blanka and Martin Valis},
	title = {Smartphone Applications Can Serve as Effective Cognitive Training Tools in Healthy Aging},
	volume = {9},
	number = {},
	pages = {436},
	year = {2018},
	doi = {10.3389/fnagi.2017.00436},
	eprint = {https://www.ncbi.nlm.nih.gov/pmc/articles/PMC5770789/},
	journal = {Frontiers in aging neuroscience}
}

@article{Turunen2019,
    doi = {10.1371/journal.pone.0219541},
    author = {Turunen, Merita AND Hokkanen, Laura AND Bäckman, Lars AND Stigsdotter-Neely, Anna AND Hänninen, Tuomo AND Paajanen, Teemu AND Soininen, Hilkka AND Kivipelto, Miia AND Ngandu, Tiia},
    journal = {PLOS ONE},
    publisher = {Public Library of Science},
    title = {Computer-based cognitive training for older adults: Determinants of adherence},
    year = {2019},
    month = {07},
    volume = {14},
    url = {https://doi.org/10.1371/journal.pone.0219541},
    pages = {1-12},
    number = {7},

}

@article{P4_medicine2013,
author = {Flores, Mauricio and Glusman, Gustavo and Brogaard, Kristin and Price, Nathan D and Hood, Leroy},
title = {P4 medicine: how systems medicine will transform the healthcare sector and society},
journal = {Personalized Medicine},
volume = {10},
number = {6},
pages = {565-576},
year = {2013},
doi = {10.2217/pme.13.57},
    note ={PMID: 25342952},

URL = { 
        https://doi.org/10.2217/pme.13.57
    
},
eprint = { 
        https://doi.org/10.2217/pme.13.57
    
}
}

@book{Reiter2000,
author = {Reiter, Ehud and Dale, Robert},
title = {Building Natural Language Generation Systems},
year = {2000},
isbn = {0521620368},
publisher = {Cambridge University Press},
address = {USA}
}

@article{PORTET2009789,
title = {Automatic generation of textual summaries from neonatal intensive care data},
journal = {Artificial Intelligence},
volume = {173},
number = {7},
pages = {789-816},
year = {2009},
issn = {0004-3702},
doi = {https://doi.org/10.1016/j.artint.2008.12.002},
url = {https://www.sciencedirect.com/science/article/pii/S0004370208002117},
author = {François Portet and Ehud Reiter and Albert Gatt and Jim Hunter and Somayajulu Sripada and Yvonne Freer and Cindy Sykes}
}

@article{Review-Assessment-Cognitive-Thought-Disorders,
title = "A Review of Automated Speech and Language Features for Assessment of Cognitive and Thought Disorders",
author = "Rohit Voleti and Liss, {Julie M.} and Visar Berisha",
year = "2020",
month = feb,
doi = "10.1109/JSTSP.2019.2952087",
language = "English (US)",
volume = "14",
pages = "282--298",
journal = "IEEE Journal on Selected Topics in Signal Processing",
issn = "1932-4553",
publisher = "Institute of Electrical and Electronics Engineers Inc.",
number = "2",
}

@inproceedings{crabbe-candito-2008-experiences,
    title = "Exp{\'e}riences d{'}analyse syntaxique statistique du fran{\c{c}}ais",
    author = "Crabb{\'e}, Beno{\^i}t  and
      Candito, Marie",
    editor = "B{\'e}chet, Fr{\'e}d{\'e}ric  and
      Bonastre, Jean-Francois",
    booktitle = "Actes de la 15{\`e}me conf{\'e}rence sur le Traitement Automatique des Langues Naturelles. Articles longs",
    month = jun,
    year = "2008",
    address = "Avignon, France",
    publisher = "ATALA",
    url = "https://aclanthology.org/2008.jeptalnrecital-long.17/",
    pages = "161--170",
    language = "fra"
}

@article{tulliani2022efficacy,
  title={Efficacy of cognitive remediation on activities of daily living in individuals with mild cognitive impairment or early-stage dementia: a systematic review and meta-analysis},
  author={Tulliani, Nikki and Bissett, Michelle and Fahey, Paul and Bye, Rosalind and Liu, Karen P. Y.},
  journal={Systematic Reviews},
  volume={11},
  number={1},
  pages={156},
  year={2022},
  doi={10.1186/s13643-022-02032-0},
  publisher={BioMed Central},
  issn={2046-4053},
  url={https://doi.org/10.1186/s13643-022-02032-0}
}

@article{BaharFuchs2019,
  author    = {Bahar-Fuchs, Alex and Martyr, Anthony and Goh, Anita M. Y. and Sabates, Julieta and Clare, Linda},
  title     = {Cognitive training for people with mild to moderate dementia},
  journal   = {Cochrane Database of Systematic Reviews},
  year      = {2019},
  volume    = {3},
  number    = {3},
  pages     = {CD013069},
  month     = {mar},
  doi       = {10.1002/14651858.CD013069.pub2},
  pmid      = {30909318},
  pmcid     = {PMC6433473},
  url       = {https://doi.org/10.1002/14651858.CD013069.pub2}
}

@article{Mowszowski_Batchelor_Naismith_2010, title={Early intervention for cognitive decline: can cognitive training be used as a selective prevention technique?}, volume={22}, DOI={10.1017/S1041610209991748}, number={4}, journal={International Psychogeriatrics}, author={Mowszowski, Loren and Batchelor, Jennifer and Naismith, Sharon L.}, year={2010}, pages={537–548}}

@inproceedings{zhou-etal-2024-psentscore-evaluating,
    title = "{PS}ent{S}core: Evaluating Sentiment Polarity in Dialogue Summarization",
    author = "Zhou, Yongxin  and
      Ringeval, Fabien  and
      Portet, Fran{\c{c}}ois",
    editor = "Calzolari, Nicoletta  and
      Kan, Min-Yen  and
      Hoste, Veronique  and
      Lenci, Alessandro  and
      Sakti, Sakriani  and
      Xue, Nianwen",
    booktitle = "Proceedings of the 2024 Joint International Conference on Computational Linguistics, Language Resources and Evaluation (LREC-COLING 2024)",
    month = may,
    year = "2024",
    address = "Torino, Italia",
    publisher = "ELRA and ICCL",
    url = "https://aclanthology.org/2024.lrec-main.1163",
    pages = "13290--13302",
}

@inproceedings{zhou-etal-2023-survey,
    title = "A Survey of Evaluation Methods of Generated Medical Textual Reports",
    author = "Zhou, Yongxin  and
      Ringeval, Fabien  and
      Portet, Fran{\c{c}}ois",
    editor = "Naumann, Tristan  and
      Ben Abacha, Asma  and
      Bethard, Steven  and
      Roberts, Kirk  and
      Rumshisky, Anna",
    booktitle = "Proceedings of the 5th Clinical Natural Language Processing Workshop",
    month = jul,
    year = "2023",
    address = "Toronto, Canada",
    publisher = "Association for Computational Linguistics",
    url = "https://aclanthology.org/2023.clinicalnlp-1.48/",
    doi = "10.18653/v1/2023.clinicalnlp-1.48",
    pages = "447--459"
}

@inproceedings{savkov-etal-2022-consultation,
    title = "Consultation Checklists: Standardising the Human Evaluation of Medical Note Generation",
    author = "Savkov, Aleksandar  and
      Moramarco, Francesco  and
      Papadopoulos Korfiatis, Alex  and
      Perera, Mark  and
      Belz, Anya  and
      Reiter, Ehud",
    editor = "Li, Yunyao  and
      Lazaridou, Angeliki",
    booktitle = "Proceedings of the 2022 Conference on Empirical Methods in Natural Language Processing: Industry Track",
    month = dec,
    year = "2022",
    address = "Abu Dhabi, UAE",
    publisher = "Association for Computational Linguistics",
    url = "https://aclanthology.org/2022.emnlp-industry.10/",
    doi = "10.18653/v1/2022.emnlp-industry.10",
    pages = "111--120"
}

@inproceedings{zhou-etal-2025-gpt,
    title = "Can {GPT} models Follow Human Summarization Guidelines? A Study for Targeted Communication Goals",
    author = "Zhou, Yongxin  and
      Ringeval, Fabien  and
      Portet, Fran{\c{c}}ois",
    editor = "Flek, Lucie  and
      Narayan, Shashi  and
      Phuong, Le Hong  and
      Pei, Jiahuan",
    booktitle = "Proceedings of the 18th International Natural Language Generation Conference",
    month = oct,
    year = "2025",
    address = "Hanoi, Vietnam",
    publisher = "Association for Computational Linguistics",
    url = "https://aclanthology.org/2025.inlg-main.17/",
    pages = "249--273"
}

@phdthesis{zhou:tel-05101086,
  TITLE = {{Affect-aware Natural Language Generation : application to dialogue and cognitive remediation session summarization in low-resource settings}},
  AUTHOR = {Zhou, Yongxin},
  URL = {https://theses.hal.science/tel-05101086},
  NUMBER = {2024GRALM063},
  SCHOOL = {{Universit{\'e} Grenoble Alpes [2020-....]}},
  YEAR = {2024},
  MONTH = Nov,
  TYPE = {Theses},
  HAL_ID = {tel-05101086},
  HAL_VERSION = {v1},
}

@inproceedings{van-der-lee-etal-2019-best,
    title = "Best practices for the human evaluation of automatically generated text",
    author = "van der Lee, Chris  and
      Gatt, Albert  and
      van Miltenburg, Emiel  and
      Wubben, Sander  and
      Krahmer, Emiel",
    editor = "van Deemter, Kees  and
      Lin, Chenghua  and
      Takamura, Hiroya",
    booktitle = "Proceedings of the 12th International Conference on Natural Language Generation",
    month = oct # "–" # nov,
    year = "2019",
    address = "Tokyo, Japan",
    publisher = "Association for Computational Linguistics",
    url = "https://aclanthology.org/W19-8643/",
    doi = "10.18653/v1/W19-8643",
    pages = "355--368"
}

@article{HUNTER2012157,
title = {Automatic generation of natural language nursing shift summaries in neonatal intensive care: BT-Nurse},
journal = {Artificial Intelligence in Medicine},
volume = {56},
number = {3},
pages = {157-172},
year = {2012},
issn = {0933-3657},
doi = {https://doi.org/10.1016/j.artmed.2012.09.002},
url = {https://www.sciencedirect.com/science/article/pii/S0933365712001170},
author = {James Hunter and Yvonne Freer and Albert Gatt and Ehud Reiter and Somayajulu Sripada and Cindy Sykes}
}

@article{10.5555/3241691.3241693,
author = {Gatt, Albert and Krahmer, Emiel},
title = {Survey of the state of the art in natural language generation: core tasks, applications and evaluation},
year = {2018},
issue_date = {January 2018},
publisher = {AI Access Foundation},
address = {El Segundo, CA, USA},
volume = {61},
number = {1},
issn = {1076-9757},
journal = {J. Artif. Int. Res.},
month = jan,
pages = {65–170},
numpages = {106}
}

@article{kueider-etal-2012-cct,
  author  = {Kueider, Alexandra M. and Parisi, Jeanine M. and Gross, Alden L. and Rebok, George W.},
  title   = {Computerized Cognitive Training with Older Adults: A Systematic Review},
  journal = {{PLOS ONE}},
  volume  = {7},
  number  = {7},
  pages   = {e40588},
  year    = {2012},
  doi     = {10.1371/journal.pone.0040588},
  pmid    = {22792378},
  pmcid   = {PMC3394709},
}

@inproceedings{reiter-sripada-2003-learning,
    title = "Learning the Meaning and Usage of Time Phrases from a Parallel Text-Data Corpus",
    author = "Reiter, Ehud  and
      Sripada, Somayajulu",
    booktitle = "Proceedings of the {HLT}-{NAACL} 2003 Workshop on Learning Word Meaning from Non-Linguistic Data",
    year = "2003",
    url = "https://aclanthology.org/W03-0611/",
    pages = "78--85"
}

@article{LYU2026104997,
title = {Natural language generation in healthcare: A review of methods and applications},
journal = {Journal of Biomedical Informatics},
volume = {176},
pages = {104997},
year = {2026},
issn = {1532-0464},
doi = {https://doi.org/10.1016/j.jbi.2026.104997},
url = {https://www.sciencedirect.com/science/article/pii/S1532046426000213},
author = {Mengxian Lyu and Xiaohan Li and Ziyi Chen and Jinqian Pan and Cheng Peng and Sankalp Talankar and Yonghui Wu}
}

@article{Lampit2014Computerized,
  author    = {A. Lampit and H. Hallock and M. Valenzuela},
  title     = {Computerized cognitive training in cognitively healthy older adults: a systematic review and meta-analysis of effect modifiers},
  journal   = {PLoS Med},
  year      = {2014},
  volume    = {11},
  number    = {11},
  pages     = {e1001756},
  month     = {Nov},
  doi       = {10.1371/journal.pmed.1001756},
  pmid      = {25405755},
  pmcid     = {PMC4236015}
}

@article{Thirunavukarasu2023Large,
  author    = {A. J. Thirunavukarasu and D. S. J. Ting and K. Elangovan and L. Gutierrez and T. F. Tan and D. S. W. Ting},
  title     = {Large language models in medicine},
  journal   = {Nat Med},
  year      = {2023},
  volume    = {29},
  number    = {8},
  pages     = {1930--1940},
  month     = {Aug},
  doi       = {10.1038/s41591-023-02448-8},
  pmid      = {37460753},
  note      = {Epub 2023 Jul 17}
}

@article{Singhal2023Large,
  author    = {K. Singhal and S. Azizi and T. Tu and S. S. Mahdavi and J. Wei and H. W. Chung and N. Scales and A. Tanwani and H. Cole-Lewis and S. Pfohl and P. Payne and M. Seneviratne and P. Gamble and C. Kelly and A. Babiker and N. Schärli and A. Chowdhery and P. Mansfield and D. Demner-Fushman and B. Agüera Y Arcas and D. Webster and G. S. Corrado and Y. Matias and K. Chou and J. Gottweis and N. Tomasev and Y. Liu and A. Rajkomar and J. Barral and C. Semturs and A. Karthikesalingam and V. Natarajan},
  title     = {Large language models encode clinical knowledge},
  journal   = {Nature},
  year      = {2023},
  volume    = {620},
  number    = {7972},
  pages     = {172--180},
  month     = {Aug},
  doi       = {10.1038/s41586-023-06291-2},
  pmid      = {37438534},
  pmcid     = {PMC10396962},
  note      = {Epub 2023 Jul 12. Erratum in: Nature. 2023 Aug;620(7973):E19. doi: 10.1038/s41586-023-06455-0}
}

@article{REITER200341,
title = {Lessons from a failure: Generating tailored smoking cessation letters},
journal = {Artificial Intelligence},
volume = {144},
number = {1},
pages = {41-58},
year = {2003},
issn = {0004-3702},
doi = {https://doi.org/10.1016/S0004-3702(02)00370-3},
url = {https://www.sciencedirect.com/science/article/pii/S0004370202003703},
author = {Ehud Reiter and Roma Robertson and Liesl M. Osman}
}

@article {Liu2023.06.28.23291916,
	author = {Liu, Chunyu and Ma, Yongpei and Kothur, Kavitha and Nikpour, Armin and Kavehei, Omid},
	title = {BioSignal Copilot: Leveraging the power of LLMs in drafting reports for biomedical signals},
	elocation-id = {2023.06.28.23291916},
	year = {2023},
	doi = {10.1101/2023.06.28.23291916},
	publisher = {Cold Spring Harbor Laboratory Press},
	URL = {https://www.medrxiv.org/content/early/2023/07/06/2023.06.28.23291916},
	journal = {medRxiv}
}

@article{10.1145/3777411,
author = {Zhang, Shengyu and Dong, Linfeng and Li, Xiaoya and Zhang, Sen and Sun, Xiaofei and Wang, Shuhe and Li, Jiwei and Hu, Runyi and Zhang, Tianwei and Wang, Guoyin and Wu, Fei},
title = {Instruction Tuning for Large Language Models: A Survey},
year = {2026},
issue_date = {May 2026},
publisher = {Association for Computing Machinery},
address = {New York, NY, USA},
volume = {58},
number = {7},
issn = {0360-0300},
url = {https://doi.org/10.1145/3777411},
doi = {10.1145/3777411},
journal = {ACM Comput. Surv.},
month = jan,
articleno = {169},
numpages = {36}
}

@inproceedings{devlin-etal-2019-bert,
    title = "{BERT}: Pre-training of Deep Bidirectional Transformers for Language Understanding",
    author = "Devlin, Jacob  and
      Chang, Ming-Wei  and
      Lee, Kenton  and
      Toutanova, Kristina",
    editor = "Burstein, Jill  and
      Doran, Christy  and
      Solorio, Thamar",
    booktitle = "Proceedings of the 2019 Conference of the North {A}merican Chapter of the Association for Computational Linguistics: Human Language Technologies, Volume 1 (Long and Short Papers)",
    month = jun,
    year = "2019",
    address = "Minneapolis, Minnesota",
    publisher = "Association for Computational Linguistics",
    url = "https://aclanthology.org/N19-1423/",
    doi = "10.18653/v1/N19-1423",
    pages = "4171--4186"
}

@inproceedings{10.5555/3495724.3496768,
author = {Baevski, Alexei and Zhou, Henry and Mohamed, Abdelrahman and Auli, Michael},
title = {wav2vec 2.0: a framework for self-supervised learning of speech representations},
year = {2020},
isbn = {9781713829546},
publisher = {Curran Associates Inc.},
address = {Red Hook, NY, USA},
booktitle = {Proceedings of the 34th International Conference on Neural Information Processing Systems},
articleno = {1044},
numpages = {12},
location = {Vancouver, BC, Canada},
series = {NIPS '20}
}

@InProceedings{pmlr-v139-radford21a,
  title = 	 {Learning Transferable Visual Models From Natural Language Supervision},
  author =       {Radford, Alec and Kim, Jong Wook and Hallacy, Chris and Ramesh, Aditya and Goh, Gabriel and Agarwal, Sandhini and Sastry, Girish and Askell, Amanda and Mishkin, Pamela and Clark, Jack and Krueger, Gretchen and Sutskever, Ilya},
  booktitle = 	 {Proceedings of the 38th International Conference on Machine Learning},
  pages = 	 {8748--8763},
  year = 	 {2021},
  editor = 	 {Meila, Marina and Zhang, Tong},
  volume = 	 {139},
  series = 	 {Proceedings of Machine Learning Research},
  month = 	 {18--24 Jul},
  publisher =    {PMLR},
  url = 	 {https://proceedings.mlr.press/v139/radford21a.html}
}

@misc{kalai2025languagemodelshallucinate,
      title={Why Language Models Hallucinate}, 
      author={Adam Tauman Kalai and Ofir Nachum and Santosh S. Vempala and Edwin Zhang},
      year={2025},
      eprint={2509.04664},
      archivePrefix={arXiv},
      primaryClass={cs.CL},
      url={https://arxiv.org/abs/2509.04664}, 
}

@inproceedings{varges-etal-2012-semscribe,
    title = "{S}em{S}cribe: Natural Language Generation for Medical Reports",
    author = "Varges, Sebastian  and
      Bieler, Heike  and
      Stede, Manfred  and
      Faulstich, Lukas C.  and
      Irsig, Kristin  and
      Atalla, Malik",
    editor = "Calzolari, Nicoletta  and
      Choukri, Khalid  and
      Declerck, Thierry  and
      Do{\u{g}}an, Mehmet U{\u{g}}ur  and
      Maegaard, Bente  and
      Mariani, Joseph  and
      Moreno, Asuncion  and
      Odijk, Jan  and
      Piperidis, Stelios",
    booktitle = "Proceedings of the Eighth International Conference on Language Resources and Evaluation ({LREC}'12)",
    month = may,
    year = "2012",
    address = "Istanbul, Turkey",
    publisher = "European Language Resources Association (ELRA)",
    url = "https://aclanthology.org/L12-1032/",
    pages = "2674--2681"
}

@inproceedings{wang-etal-2025-towards-adapting,
    title = "Towards Adapting Open-Source Large Language Models for Expert-Level Clinical Note Generation",
    author = "Wang, Hanyin  and
      Gao, Chufan  and
      Liu, Bolun  and
      Xu, Qiping  and
      Hussein, Guleid  and
      Labban, Mohamad El  and
      Iheasirim, Kingsley  and
      Korsapati, Hariprasad Reddy  and
      Outcalt, Chuck  and
      Sun, Jimeng",
    editor = "Che, Wanxiang  and
      Nabende, Joyce  and
      Shutova, Ekaterina  and
      Pilehvar, Mohammad Taher",
    booktitle = "Findings of the Association for Computational Linguistics: ACL 2025",
    month = jul,
    year = "2025",
    address = "Vienna, Austria",
    publisher = "Association for Computational Linguistics",
    url = "https://aclanthology.org/2025.findings-acl.626/",
    doi = "10.18653/v1/2025.findings-acl.626",
    pages = "12084--12117",
    ISBN = "979-8-89176-256-5"
}

@Article{Bedi2026,
author={Bedi, Suhana and Cui, Hejie and Fuentes, Miguel and Unell, Alyssa and Wornow, Michael and Banda, Juan M. and Kotecha, Nikesh and Keyes, Timothy and Mai, Yifan and Oez, Mert and Qiu, Hao and Jain, Shrey and Schettini, Leonardo and Kashyap, Mehr and Fries, Jason Alan and Swaminathan, Akshay and Chung, Philip and Haredasht, Fateme Nateghi and Lopez, Ivan and Aali, Asad and Tse, Gabriel and Nayak, Ashwin and Vedak, Shivam and Jain, Sneha S. and Patel, Birju and Fayanju, Oluseyi and Shah, Shreya and Goh, Ethan and Yao, Dong-han and Soetikno, Brian and Reis, Eduardo and Gatidis, Sergios and Divi, Vasu and Capasso, Robson and Saralkar, Rachna and Chiang, Chia-Chun and Jindal, Jenelle and Pham, Tho and Ghoddusi, Faraz and Lin, Steven and Chiou, Albert S. and Hong, Hyo Jung and Roy, Mohana and Gensheimer, Michael F. and Patel, Hinesh and Schulman, Kevin and Dash, Dev and Char, Danton and Downing, Lance and Grolleau, Francois and Black, Kameron and Mieso, Bethel and Zahedivash, Aydin and Yim, Wen-wai and Sharma, Harshita and Lee, Tony and Kirsch, Hannah and Lee, Jennifer and Ambers, Nerissa and Lugtu, Carlene and Sharma, Aditya and Mawji, Bilal and Alekseyev, Alex and Zhou, Vicky and Kakkar, Vikas and Helzer, Jarrod and Revri, Anurang and Bannett, Yair and Daneshjou, Roxana and Chen, Jonathan and Alsentzer, Emily and Morse, Keith and Ravi, Nirmal and Aghaeepour, Nima and Kennedy, Vanessa and Chaudhari, Akshay and Wang, Thomas and Koyejo, Sanmi and Lungren, Matthew P. and Horvitz, Eric and Liang, Percy and Pfeffer, Michael A. and Shah, Nigam H.},
title={Holistic evaluation of large language models for medical tasks with MedHELM},
journal={Nature Medicine},
year={2026},
month={Mar},
day={01},
volume={32},
number={3},
pages={943-951},
issn={1546-170X},
doi={10.1038/s41591-025-04151-2},
url={https://doi.org/10.1038/s41591-025-04151-2}
}

@Inbook{Pauws2019,
author="Pauws, Steffen
and Gatt, Albert
and Krahmer, Emiel
and Reiter, Ehud",
title="Making Effective Use of Healthcare Data Using Data-to-Text Technology",
bookTitle="Data Science for Healthcare: Methodologies and Applications",
year="2019",
publisher="Springer International Publishing",
address="Cham",
pages="119--145",
isbn="978-3-030-05249-2",
doi="10.1007/978-3-030-05249-2_4",
url="https://doi.org/10.1007/978-3-030-05249-2_4"
}

\appendix

\section{Cognitive Training Exercises}
\label{appendix:theradia_exercises} 

Eight exercises were proposed for cognitive training during the THERADIA Wizard-of-Oz experiments:

\begin{enumerate}[noitemsep,topsep=2pt]
    \item Exo 1 \textbf{``Retrouvez votre chemin'' [Find your way]}
    \item Exo 2 \textbf{``Objets où êtes-vous ?'' [Objects, Where are You?]}
    \item Exo 3 \textbf{``Que d’accros'' [This Story is Full of Blanks]}
    \item Exo 4 \textbf{``Jeux de blasons'' [Blazon game]}
    \item Exo 5 \textbf{``Mettez de l’ordre dans ces comptes'' [The Right Count]}
    \item Exo 6 \textbf{``Garçon SVP'' [Restaurant]}
    \item Exo 7 \textbf{``Menez l'enquête'' [Carry out the investigation]}
    \item Exo 8 \textbf{``Tour Hanoï'' [Towers of Hanoi]}
\end{enumerate}
    
\begin{figure*}[!ht]
	   \begin{subfigure}{\linewidth}
             \centering
		\includegraphics[width=.8\textwidth]{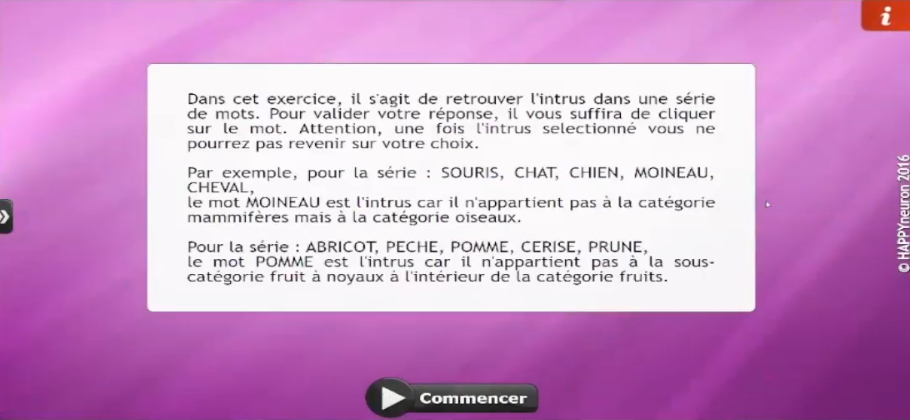}
		\caption{Instruction}
		\label{fig:exo7_subfig1}
	   \end{subfigure}
	\vfill
	     \begin{subfigure}{\linewidth}
               \centering
		 \includegraphics[width=.8\textwidth]{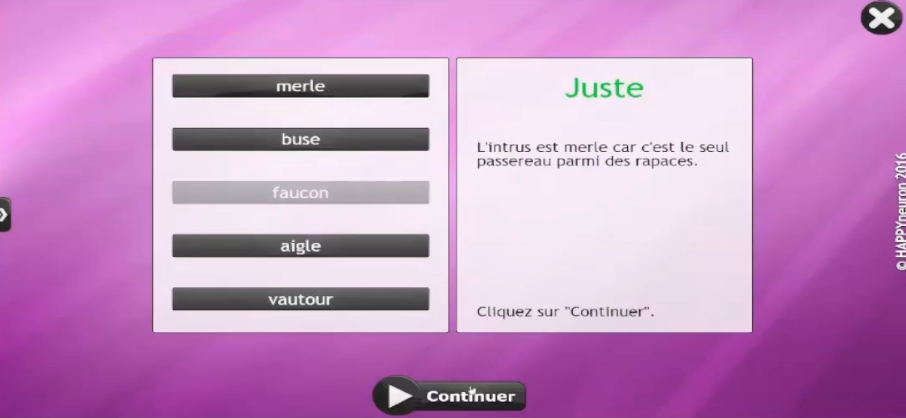}
		 \caption{Exercise}
		 \label{fig:exo7_subfig2}
	      \end{subfigure}
	\caption{Screenshot of \textit{Exo 7 Menez l’enquête} interface.}
	\label{fig:exo7}
\end{figure*}

As described in Section~\ref{sec:theradia_report_form}, sessions for young and senior participants comprised eight exercises, each repeated once. For MCI participants, the number of exercises was reduced to four (Exo 1, 2, 3, and 7), each repeated once, to accommodate the difficulties of their involvement. Instructions were provided before each exercise; for example, Exo 7 opens with instructions and examples for identifying the odd word in a series (Figure~\ref{fig:exo7_subfig1}).

\textbf{Exo 1 Retrouvez votre chemin}

\begin{figure*}[!ht]
    \centering
    \includegraphics[width=.8\textwidth]{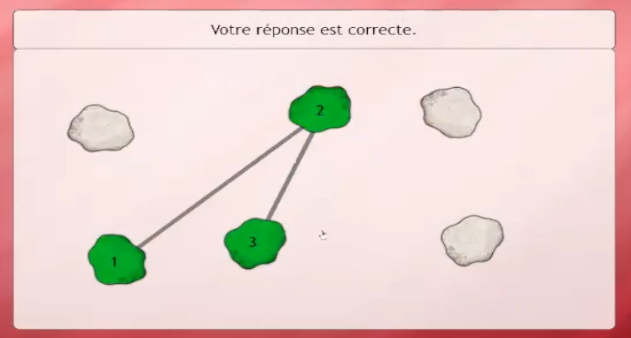}
    \caption{Screenshot of \textit{Exo 1 Retrouvez votre chemin} interface.}
    \label{fig:exo1}
\end{figure*}

Shown in Figure~\ref{fig:exo1}, this exercise trains visuo-spatial memory, defined as the ability to remember the position of elements in one's surroundings. Participants must memorize the order in which stones appear and click on them in the correct sequence.

\textbf{Exo 2 Objets où êtes-vous ?}

\begin{figure*}[!ht]
    \centering
    \includegraphics[width=.8\textwidth]{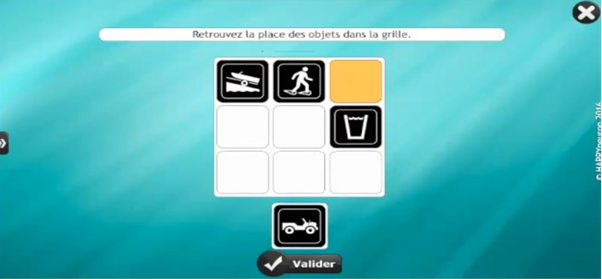}
    \caption{Screenshot of \textit{Exo 2 Objets où êtes-vous ?} interface.}
    \label{fig:exo2}
\end{figure*}

Shown in Figure~\ref{fig:exo2}, this exercise presents a sequence of images appearing one by one in a grid. Participants have six seconds to memorize each image and its location, then must recall and click on the correct grid positions. At higher difficulty levels, participants must memorize and recall five images rather than four.

\textbf{Exo 3 Que\_d’accros}

\begin{figure*}[!ht]
    \centering
    \includegraphics[width=.8\textwidth]{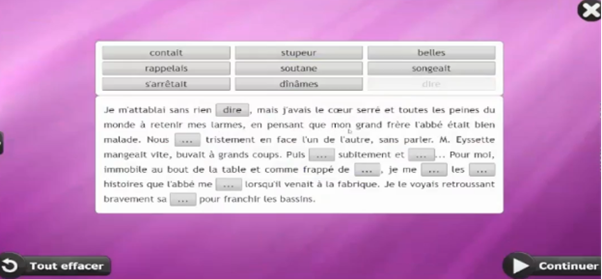}
    \caption{Screenshot of \textit{Exo 3 Que\_d’accros} interface.}
    \label{fig:exo3}
\end{figure*}

Shown in Figure~\ref{fig:exo3}, this exercise tests knowledge of the French language and reasoning skills. Participants are asked to reconstruct a text from which certain words have been removed.

\textbf{Exo 4 Jeux de blasons}

\begin{figure*}[!ht]
    \centering
    \includegraphics[width=.8\textwidth]{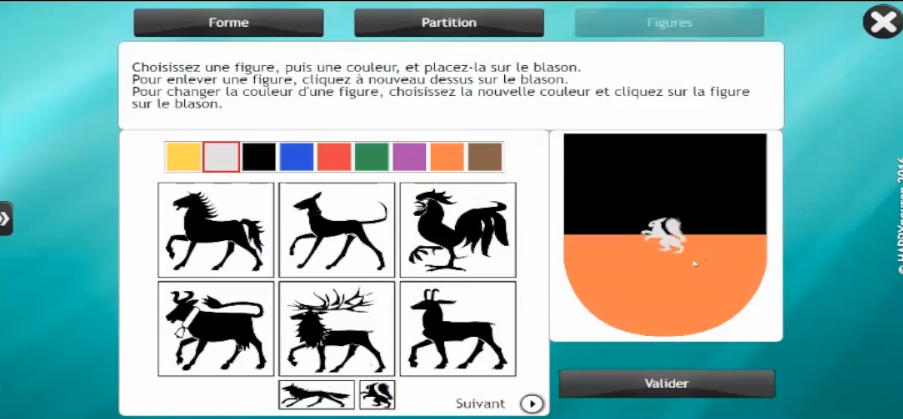}
    \caption{Screenshot of \textit{Exo 4 Jeux de blasons} interface.}
    \label{fig:exo4}
\end{figure*}

Shown in Figure~\ref{fig:exo4}, this exercise trains visual memory and attention. Participants memorize a coat of arms, complete a distracting task, then reconstruct the coat of arms by selecting its constituent elements and colors: (1) the shield shape, (2) the two shield colors, and (3) the shape, position, and color of the central figure.

\textbf{Exo5 Mettez de l’ordre dans ces comptes}

\begin{figure*}[!ht]
    \centering
    \includegraphics[width=.8\textwidth]{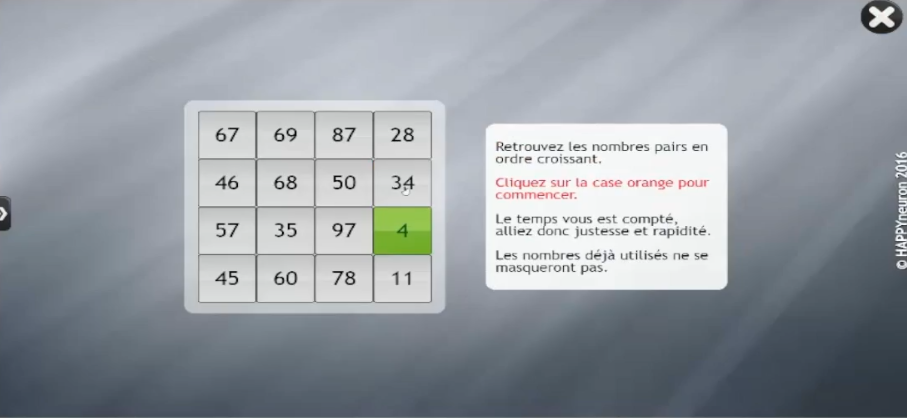}
    \caption{Screenshot of \textit{Exo5 Mettez de l’ordre dans ces comptes} interface.}
    \label{fig:exo5}
\end{figure*}

Shown in Figure~\ref{fig:exo5}, this exercise targets calculation, number manipulation, and reasoning, with speed and accuracy both essential. Participants sort numbers in ascending or descending order, restricted to odd or even numbers as instructed, across multiple trials with varying instructions. The goal is to respond as quickly and accurately as possible.

\textbf{Exo 6 Garçon SVP}

\begin{figure*}[!ht]
      \centering
	   \begin{subfigure}{\linewidth}
             \centering
		\includegraphics[width=.8\textwidth]{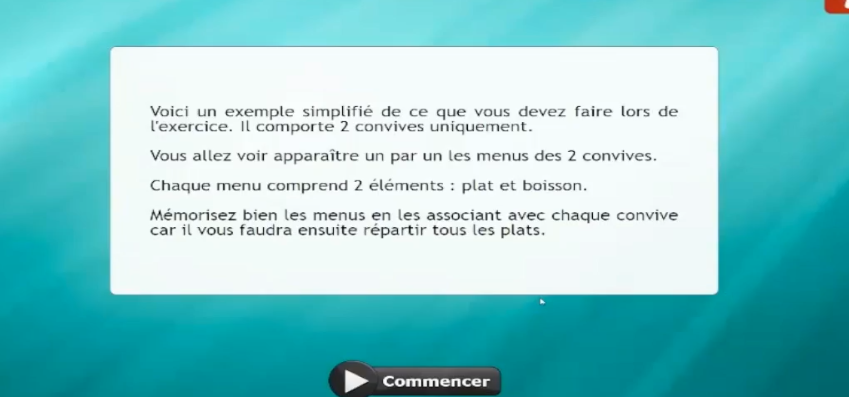}
		\caption{Instruction}
		\label{fig:subfig1}
	   \end{subfigure}
	\vfill
	     \begin{subfigure}{\linewidth}
               \centering
		 \includegraphics[width=.8\textwidth]{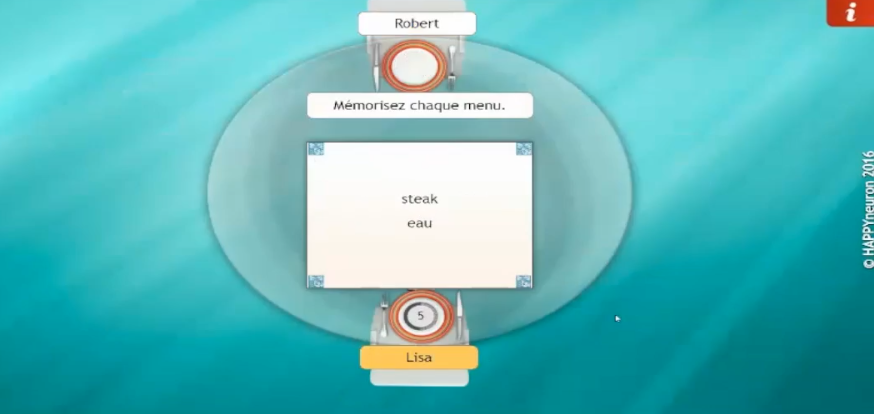}
		 \caption{Exercise}
		 \label{fig:subfig3}
	      \end{subfigure}
	\caption{Screenshot of \textit{Exo 6 Garçon SVP} interface.}
	\label{fig:exo6}
\end{figure*}

Shown in Figure~\ref{fig:exo6}, this exercise trains both visual and verbal memory. Participants take on the role of a restaurant waiter and must memorize customers' orders within a limited time per order, then click on the correct customer to serve each dish. At higher difficulty levels, each order contains three items.

\textbf{Exo 7 Menez l’enquête}

Shown in Figure~\ref{fig:exo7}, this exercise requires participants to identify the odd word in a series, drawing on French language knowledge. This skill supports speaking, reading, and lexical retrieval in everyday situations. Participants click on the word they believe does not belong; once selected, the answer cannot be changed. For example, in the series ``Mouse, Cat, Dog, Sparrow, Horse'', the correct answer is ``Sparrow'' as it is a bird while the others are mammals.

\textbf{Exo 8 Tour Hanoï}

\begin{figure*}[!ht]
    \centering
    \includegraphics[width=.8\textwidth]{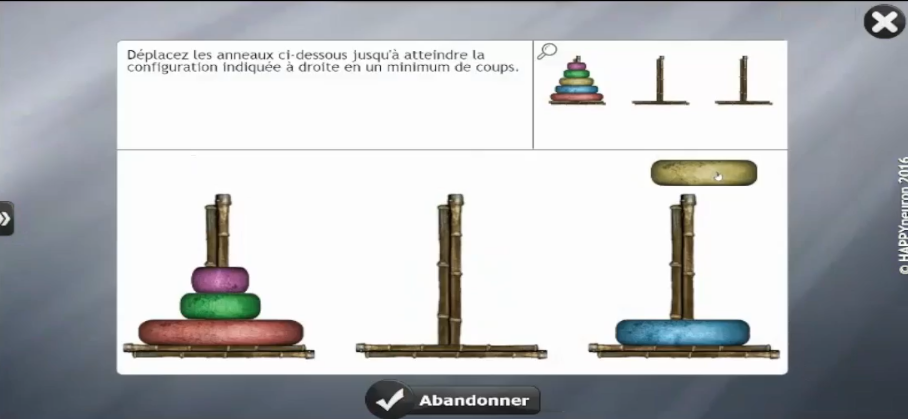}
    \caption{Screenshot of \textit{Exo 8 Tour Hanoï} interface.}
    \label{fig:exo8}
\end{figure*}

Shown in Figure~\ref{fig:exo8}, this exercise stimulates reasoning and planning skills, requiring participants to organize, prioritize, and sequence steps toward a goal. Participants must find a solution within 10 to 14 moves and are given three attempts, with no time limit, encouraging strategic thinking.

\section{THERADIA Dataset}

\subsection{Statistics}
\label{appendix:theradia_statistics}

Table~\ref{tab:tbl_theradia_data_num} presents statistics for the THERADIA dataset. For young and senior groups, participants are divided into five subgroups (A--E) according to whether they received an induction or not. Induction refers to the deliberate variation of exercise difficulty across repetitions, designed to elicit emotions in participants without their awareness. In the non-induction subgroup (\textit{E}), exercise difficulty remains constant throughout the session.

\begin{table*}[ht]
  \centering
  \footnotesize
    \caption{THERADIA dataset statistics (updated April 5, 2024). Young and senior participants are divided into subgroups \textit{A--E}, with sessions lasting 1--1.5 hours. MCI participants completed shorter sessions of 30--45 minutes.}
  \begin{tabular}{l|p{0.25\linewidth}|l|l|p{0.2\linewidth}}
    \toprule
      & Group & Folder        & N° participants & N° sessions for each participant \\
    \midrule
      & Elder participants (A-participants)     & data\_agés\_A     & 8  & 1   \\
      &         & data\_agés\_B     & 8   & 1    \\
      &         & data\_agés\_C     & 8   & 1     \\
      &         & data\_agés\_D     & 8   & 1     \\
      &         & data\_agés\_E     & 20 & 2  \\
    \midrule
    & Young participants (J-participants)  & data\_jeunes\_A   & 8   & 1  \\
      &         & data\_jeunes\_B   & 8  & 1 or 2     \\
      &         & data\_jeunes\_C   & 8     & 1 or 2  \\
      &         & data\_jeunes\_D   & 8  & 1 or 2   \\
      &         & data\_jeunes\_E   & 20  & 2   \\
    \midrule 
      & Participants with early-stage Alzheimer’s disease (MCI)     & M-participants & 9  & 1 \\
    \midrule
Total &         &                   & 113            &   \\       
\bottomrule
\end{tabular}

  \label{tab:tbl_theradia_data_num}
\end{table*}

\subsection{Session Data Illustrations}
\label{appendix:theradia_data_examples}

\begin{figure*}[ht]
    \centering
    \includegraphics[
    trim=0mm 60mm 0mm 60mm,
    clip,
    width=\linewidth
    ]{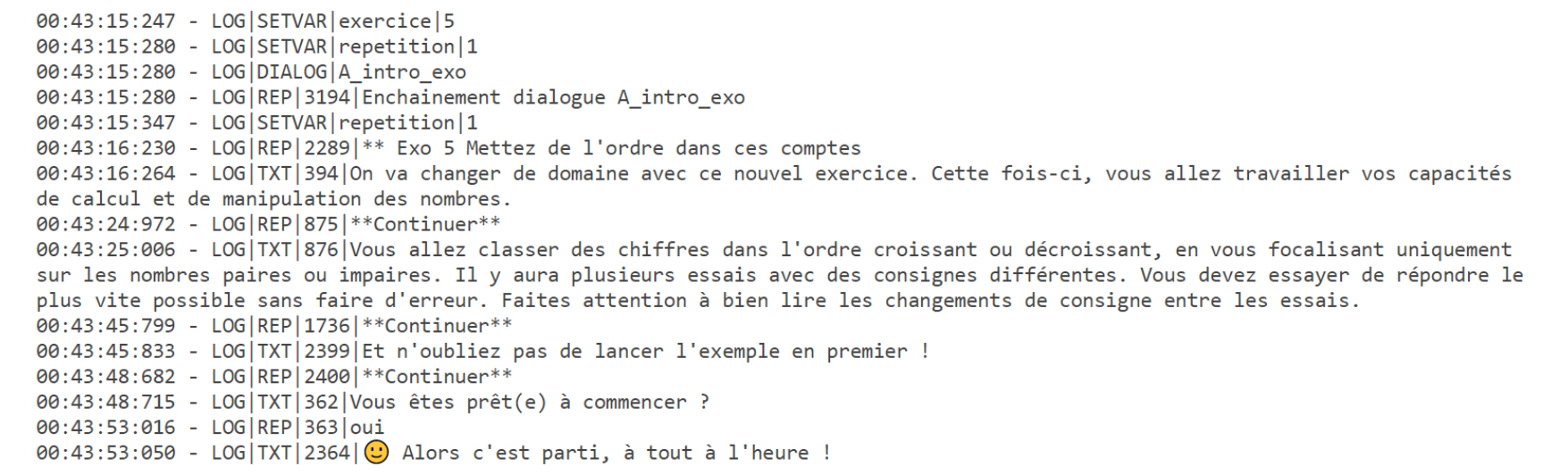}
    \caption{Excerpt from the log file recorded while participant \textit{A01A} performed the first repetition of exercise \textit{Exo 5}.
    \textit{LOG|REP}: Button clicked by the Wizard (the person controlling the virtual assistant) summarizing the comments of the participant,
    \textit{LOG|TXT}: Reply read by the Wizard, \textit{LOG|ENDGAME}: Exercise result, \textit{CONF|DIALOG}: Full-screen display of avatar, \textit{LOG|SETVAR}: Registering a variable.}
    \label{fig:log_excerpt}
\end{figure*}

Figure~\ref{fig:log_excerpt} presents an excerpt from the log file recorded during participant A01A's first repetition of exercise Exo~5.

\section{Session Report Structure}

\subsection{Reporting Requirements}

Table~\ref{table:variables_list_therapist} lists the essential information elements for remote clinical observation reports, as identified in Section~\ref{subsubsec:remote_variables} through expert consultation with a collaborating speech therapist.

\begin{table*}[ht]
\centering
\footnotesize
\caption{Essential information to communicate in a remediation session report, as proposed by an expert speech therapist.}
\begin{tabular}{p{3.5cm}|p{9cm}}
\toprule
\textbf{Variables} & \textbf{Examples}\\
\midrule
1- Sleep / Drowsiness & Narcolepsy \\ \midrule
2- Visual gene & Gets too close to the screen - rubs eyes\\ \midrule
3- Auditory gene & Significantly raises the volume - expresses discomfort in discriminating the text heard\\ \midrule
4- Difficulty in understanding & Of the text read or heard - if expressed \\ \midrule
5- Agitation & Behavioral or Psychomotor\\ \midrule
6- Network problem & Connection/disconnection \\ \midrule
7- Initiate communication with the avatar & What kind of communication? Sharing experiences, questions? What kind of questions? \\ \midrule
8- Questions asked by the avatar before starting the training & How does the patient feel \\ \midrule
9- Question asked by the avatar during training & For failed or very slow exercises - Did you like this exercise? \\ \midrule
10- Questions asked by the avatar at the end of the session & - Did you encounter any particular difficulties? - Are you satisfied with your performance? \\ \midrule
11- Information on the patient's behavior specific to the training: & - Look at the example
- Delays between instructions and answers, uses or not the time limit 
- Shifts his/her attention (answers the phone, looks away, talks to someone, moves during the session).
- Manipulation difficulties: does not click in the right place, has difficulty using the mouse
- Reactions to the avatar's or helper's stimuli: adapted or not, changes his/her
behavior.
- Consultation of solutions or not, delays between the end of an exercise and the next one.
- Attempts to adapt to the difficulties encountered: verbalizes strategies, looks for external help
(notes, asks the caregiver)
- Manifestations of the emotional state: verbalizations (questioning, satisfaction, disappointment,
concern...), gestures and mimics (breath, smiles,).\\ 
\midrule
12- Regularity, involvement of the patient & \% of sessions carried out, periods of interruption of the trainings\\ 
\midrule
13- Dates and times of the sessions carried out & Correlation between results and days, times.\\ \midrule
14- Context of the training, environment & Alone or in a common room, occurrence of external interruptions during sessions (visits, phone calls, pain...) \\ 
\midrule
15- Presence of a third party and type of intervention & -Verbal: practical advice, gives answers
encourages, reminds of the instructions, \\
& -Gestural: manipulates designates elements on the screen \\ 
\bottomrule
\end{tabular}
\label{table:variables_list_therapist}
\end{table*}

\subsection{Clinical Notes Vocabulary}

Table~\ref{tab:theradia_note_vocabulary} presents the vocabulary taxonomy derived from analysis of speech therapists' clinical notes (Section~\ref{subsubsec:analysis_real_notes}), organized into ten categories: comprehension, production, emotion, execution, attention, behavior, motivation, memory, reasoning, and self-evaluation.

\begin{table*}[!ht]
\centering
\footnotesize
\caption{Vocabulary identified in speech therapists' notes, with definitions.}
   \begin{tabular}{p{2cm}|p{3.5cm}|p{9.5cm}}
\toprule
\textbf{Category} & \textbf{Elements / Vocabulary} & \textbf{Definition} \\
\midrule
    \textbf{Comprehension} & Oral comprehension & How the patient understands in exchanges, answers simple questions in a particular context $\rightarrow$ comprehension in discussions; 
    
    Show correct pictures in front of them $\rightarrow$ comprehension of words/phrases. \\
\midrule
    \textbf{Production} & Lexicon & The patient can access the right words correctly if he has an image in his head but can't say it. Difficulty depending on complaint/pathology. \\
    & Semantics & Semantic lexicon, find word meanings, what does it mean. \\
    & Syntax & Sentence construction or understanding. \\  
    & Informativeness & Unclear messages, the patient answers the question but not on the point, e.g.: "Tell us about your day." => the patient changes the subject, talk nonsense, sometimes has trouble remembering things too. \\
     & Phonological (difficulties) & Phonological errors, e.g. \textit{loupe} $\leftarrow$ \textit{louche}. \\
    & Periphrases & Use descriptions to describe a word (saying the thing but without using the right words). \\
\midrule
    \textbf{Emotion} & Psycho-affective state & The patient is tired, happy or not. \\
\midrule
    \textbf{Execution} & Dysexecutive syndrome & “Too many tests” (conclusion of an assessment), attention, concentration. E.g. How the patient organizes himself to go to the maze connect in alphabetical order - letter/number”. \\
    & (Stagnating) performance & From one treatment plan to the next, the patient has the same result as before. \\ 
    \midrule
    \textbf{Attention} & Attention labile & Someone has trouble staying focused on a task/exercise. \\
    & Disruption of selective and divided attention & Selective: concentrating on one thing we're talking about; Divided: able to do two things at once. \\
    & (Ability to) concentrate (very fluctuating) & Sometimes the patient can concentrate for 10 minutes without a problem, sometimes after two minutes he has stalled. \\
    & Attentional resources & If the patient is attentive, he has no trouble finding the right words; otherwise, for example when his son paces the room, the patient loses track and makes mistakes. \\
\midrule
    \textbf{Behavior} & Praxis disorders & Not good movements such as clicks buttons, opens pen, etc. \\
    & Initiation + anticipation & Initiation: given the instructions but the patient doesn't do it, doesn't start, doesn't begin. 
    
    Anticipation/planning: in a labyrinth, we usually plan the right paths in advance, but patients often go right up to the wall before realizing, “ah, it's not working”. \\
\midrule
    \textbf{Motivation} & Voluntary & Different from the attentive (focused): the patient is motivated, even when faced with difficulties.\\
    & Eager to understand & Desire to understand. There are patients who want to understand the tests, the treatment plan, etc. \\
\midrule
    \textbf{Memory} & Mnesic and gnosic disorders & Mnesic: memory; Gnosic: recognizing noisy stimuli, images, and objects. \\
    & Forgetting as you go along & Equivalent to a memory (mnésique) disorder (general theme), use "forgetting" instead. As you go along, you forget what was done ten seconds or ten minutes before. \\
\midrule
    \textbf{Reasoning} & (Pertinent) remarks & The patient asks: “Are these exercises proposed to see if I have a good memory?" The patient sees that he is in difficulty, and understands everything that has happened. \\
    & (Ability to) reflect & Reason logically, to think holistically about life. \\
    & Put one's thoughts in order & Whatever is logical, with order; = dysexecutive.
    Sometimes the patient is with attention and reasoning, but can't organize, e.g. say everything in the wrong order. \\        
\midrule
    \textbf{Self-evaluation} & Self-evaluation & The patient's estimated affective state; how he feels about his performance. \\
\bottomrule
\end{tabular}
\label{tab:theradia_note_vocabulary}
\end{table*}

\subsection{Initial Taxonomy of Variables}
\label{appendix:theradia_taxonomy}

Table~\ref{tab:theradia_taxonomy} presents the complete taxonomy derived from the two sources described in Section~\ref{sec:theradia_report_form}: remote clinical observation variables and vocabulary extracted from speech therapists' notes.

\begin{table*}[ht]
  \centering
  \scriptsize 
    \caption{Combined taxonomy derived from (1) information types identified for remote clinical observation (highlighted in light gray) and (2) keywords and definitions from speech therapists' clinical notes.}
  \begin{tabular}{p{1.5cm}|p{3cm}|p{10.5cm}}
\toprule
\textbf{Category} & \textbf{Elements/Vocabulary} & \textbf{Definition} \\
\hline
    \textbf{Comprehension} & Oral comprehension & How the patient understands in exchanges, answers simple questions in a particular context $\rightarrow$ comprehension in discussions;

    Show correct pictures in front of them $\rightarrow$ comprehension of words/phrases. \\
\hline
    \textbf{Production} & Lexicon & The patient can access the right words correctly if he has an image in his head but can't say it. Difficulty depending on complaint/pathology. \\
    & Semantics & Semantic lexicon, find word meanings, what does it mean. \\
    & Syntax & Sentence construction or understanding. \\  
    & Informativeness & Unclear messages, the patient answers the question but not on the point, e.g.: "Tell us about your day." => the patient changes the subject, talk nonsense, sometimes has trouble remembering things too. \\
     & Phonological (difficulties) & Phonological errors, e.g. \textit{loupe} $\leftarrow$ \textit{louche}. \\
    & Periphrases & Use descriptions to describe a word (saying the thing but without using the right words). \\
\hline

    \rowcolor{lightgray}        
    \textbf{Communication} & Communication and questions & Patient-initiated communication with the avatar.
Avatar's questions before, during, and after training. \\
\hline

    \textbf{Emotion} & Psycho-affective state & The patient is tired, happy or not. \\
\hline
    \textbf{Execution} & Dysexecutive syndrome & “Too many tests” (conclusion of an assessment), attention, concentration. E.g. How the patient organizes himself to go to the maze connect in alphabetical order - letter/number”. \\
    & (Stagnating) performance & From one treatment plan to the next, the patient has the same result as before. \\ 

        \rowcolor{lightgray}
    & Regularity, patient involvement & \% of sessions completed, periods of training interruption \\
    \rowcolor{lightgray}
    & Dates and hours of sessions completed & Correlation of results with days and hours. Example: “Out of a 45-minute session - he only completed 6 activities today - This score is down on the 12 previous sessions when he managed to complete 10 activities in 45 minutes.” \\
    \rowcolor{lightgray}
    & Training context, environment & Alone or in a shared room, occurrence of external stimuli during sessions (visits, telephone, pain...) \\
    \rowcolor{lightgray}
    & Presence of a third party and type of intervention & - Verbal: practical advice, gives answers, encourages, reminds of instructions; - Gestural: designates elements on the screen. \\
    \rowcolor{lightgray}
    & Network problems & Connection / disconnection  \\
    \hline

    \textbf{Attention} & Attention labile & Someone has trouble staying focused on a task/exercise. \\
    & Disruption of selective and divided attention & Selective: concentrating on one thing we're talking about; Divided: able to do two things at once. \\
    & (Ability to) concentrate (very fluctuating) & Sometimes the patient can concentrate for 10 minutes without a problem, sometimes after two minutes he has stalled. \\
    & Attentional resources & If the patient is attentive, he has no trouble finding the right words; otherwise, for example, when his son paces the room, the patient loses track and makes mistakes. \\
\hline
    \textbf{Behavior} & Praxis disorders & Not good movements such as clicks buttons, opens pen, etc. \\
    & Initiation + anticipation & Initiation: given the instructions but the patient doesn't do it, doesn't start, doesn't begin. 
    
    Anticipation/planning: in a labyrinth, we usually plan the right paths in advance, but patients often go right up to the wall before realizing, “ah, it's not working”. \\

    \rowcolor{lightgray}
    & Agitation & Behavioral or psychomotor \\
    \rowcolor{lightgray}
    & Information on patient behavior specific to training & - Consults the example 
    
    - Delays between instructions and answers, use or not the allotted time 
    
    - Shifts attention (answers phone, looks away, talks to someone, moves during session). 
    
    - Manipulation difficulties: doesn't click in the right place, has difficulty using the mouse. 
    
    - Reactions to stimuli from the avatar or caregiver: adapted responses or not, changes in behavior. 
    
    - Consultation of solutions or not, delays between the end of one exercise and the next. 
    
    - Attempts to adapt to difficulties encountered: verbalizes strategies, seeks external help (notes, calls on caregiver) 
    
    - Displays of emotional state: verbalizations (questioning, satisfaction, disappointment, concern, etc.), gestures and facial expressions (breathing, smiles, etc.).  \\
    \rowcolor{lightgray}
    & Sleep / drowsiness & Narcolepsy   \\
    \rowcolor{lightgray}
    & Visual discomfort & E.g. The patient gets too close to the screen - rubs his eyes. \\    
    \rowcolor{lightgray}
    & Auditory discomfort & Significantly increases the volume - expresses discomfort in discriminating the text heard. \\
    \hline
    
    \textbf{Motivation} & Voluntary & Different from the attentive (focused): the patient is motivated, even when faced with difficulties.\\
    & Eager to understand & Desire to understand. There are patients who want to understand the tests, the treatment plan, etc. \\
\hline
    \textbf{Memory} & Mnesic and gnosic disorders & Mnesic: memory; Gnosic: recognizing noisy stimuli, images, and objects. \\
    & Forgetting as you go along & Equivalent to a memory (mnésique) disorder (general theme), use "forgetting" instead. As you go along, you forget what was done ten seconds or ten minutes before. \\
\hline
    \textbf{Reasoning} & (Pertinent) remarks & The patient asks: “Are these exercises proposed to see if I have a good memory?" The patient sees that he is in difficulty, and understands everything that has happened. \\
    & (Ability to) reflect & Reason logically, to think holistically about life. \\
    & Put one's thoughts in order & Whatever is logical, with order; = dysexecutive.
    Sometimes the patient is with attention and reasoning, but can't organize, e.g. say everything in the wrong order. \\        
\hline
    \textbf{Self-evaluation} & Self-evaluation & The patient's estimated affective state; how he feels about his performance. \\
\bottomrule
\end{tabular}

  \label{tab:theradia_taxonomy}
\end{table*}

\section{Note on Linguistic Descriptors}
\label{appendix:theradia_linguistic}

\subsection{Feature Extraction}

To calculate lexical density, we applied part-of-speech (POS) tagging \citep{crabbe-candito-2008-experiences}, which assigns grammatical labels (part of speech, case, tense, etc.) to each token in a text. We used the \texttt{french-camembert-postag-model}\footnote{\url{https://huggingface.co/gilf/french-camembert-postag-model}; the base tokenizer and model is \texttt{camembert-base}.}, a POS tagging model for French trained on the publicly available \textit{free-french-treebank} dataset.

\subsection{Preprocessing Dialogue Transcripts}

The CSV transcription files contain diacritics marking non-verbal communication, including prosodic, facial, and postural expressions that can span several seconds. These must be excluded from linguistic analysis (e.g., speech duration and speech rate calculations). The diacritics used in the transcription conventions are as follows:

\begin{itemize}[noitemsep,topsep=2pt]
    \item \verb|<?>| $\Rightarrow$ Used when segmentation or transcription is uncertain; such cases are reviewed collectively on a weekly basis.
    \item \verb|<di>| $\Rightarrow$ Placed immediately before the first utterance directed at the virtual assistant (start of interaction).
    \item \verb|<fi>| $\Rightarrow$ Placed immediately after the last utterance directed at the virtual assistant (end of interaction).
    \item \verb|<nv>| $\Rightarrow$ Used to mark non-verbal utterances that are distinct from verbal ones.
    \item Other symbols (e.g., <ii> <ri>) $\Rightarrow$ Used within an utterance to mark non-verbal expressions (affective or otherwise). 
\end{itemize} 

During preprocessing, any sequence consisting solely of a diacritic is removed from the analysis, regardless of its duration or token count.

\subsection{Norms for Linguistic Indicators}

The linguistic indicator norms used for comparison in the second report table are derived from 39 sessions involving 20 senior participants from subgroup E (participants who received no induction; see Appendix~\ref{appendix:theradia_statistics}). Norms are computed as the median, first quartile, and third quartile of each indicator, providing a robust reference for assessing the linguistic profile of a given session.

\end{document}